\pdfoutput=1

\documentclass[11pt]{article}

\usepackage{acl}
\usepackage{booktabs}

\usepackage{times}
\usepackage{latexsym}
\usepackage{comment}
\usepackage{multirow}
\usepackage{subfigure}
\usepackage[T1]{fontenc}

\usepackage[utf8]{inputenc}

\usepackage{microtype}

\usepackage{inconsolata}

\usepackage{graphicx}
\usepackage{colortbl}
\usepackage{color}
\usepackage{xcolor}
\usepackage[ruled, linesnumbered]{algorithm2e}
\usepackage{amsmath}
\usepackage{fontawesome}
\usepackage{pifont}
\usepackage{amssymb}
\usepackage{utfsym}

\title{WebUIBench: A Comprehensive Benchmark for Evaluating Multimodal Large Language Models in WebUI-to-Code }

\author{
 \textbf{Zhiyu Lin\textsuperscript{1,3}\thanks{Equal Contribution}\thanks{Work done during an internship at TeleAI}},
 \textbf{Zhengda Zhou\textsuperscript{1,4}\footnotemark[1]\footnotemark[2]},
 \textbf{Zhiyuan Zhao\textsuperscript{1}\footnotemark[1]},
 \\
 \textbf{Tianrui Wan\textsuperscript{2}},
 \textbf{Yilun Ma\textsuperscript{2}},
 \textbf{Junyu Gao\textsuperscript{1,2}\thanks{Corresponding Authors: X. Li and J. Gao}},
 \textbf{Xuelong Li\textsuperscript{1}\footnotemark[3]}
\\
\small{
 \textsuperscript{1}Institute of Artificial Intelligence (TeleAI), China Telecom,}
\\
\small{
 \textsuperscript{2}Northwestern Polytechnical University,}
\small{
 \textsuperscript{3}Beijing Jiaotong University,}
\small{
 \textsuperscript{4}Nanjing University}
\\
\small{zyllin@bjtu.edu.cn; zhengdazhou@smail.nju.edu.cn; tuzixini@gmail.com}
\\
\small{tianrui@mail.nwpu.edu.cn; mayilun@mail.nwpu.edu.cn; gjy3035@gmail.com; xuelong\_li@ieee.org}
\\
\faGithub: \small\url{https://github.com/MAIL-Tele-AI/WebUIBench}
}

\begin{document}
\maketitle


\begin{abstract}


With the rapid advancement of Generative AI technology, Multimodal Large Language Models(MLLMs) have the potential to act as AI software engineers capable of executing complex web application development. Considering that the model requires a confluence of multidimensional sub-capabilities to address the challenges of various development phases, constructing a multi-view evaluation framework is crucial for accurately guiding the enhancement of  development efficiency. However, existing benchmarks usually fail to provide an assessment of sub-capabilities and focus solely on webpage generation outcomes. In this work, we draw inspiration from the principles of software engineering and further propose WebUIBench, a benchmark systematically designed to evaluate MLLMs in four key areas: \textit{WebUI Perception}, \textit{HTML Programming}, \textit{WebUI-HTML Understanding}, and \textit{WebUI-to-Code}. WebUIBench comprises 21K high-quality question-answer pairs derived from over 0.7K real-world websites. The extensive evaluation of 29 mainstream MLLMs uncovers the skill characteristics and various weakness that models encountered during the development process.

\end{abstract}

\begin{figure}
    \centering
    \includegraphics[width=1.0\linewidth]{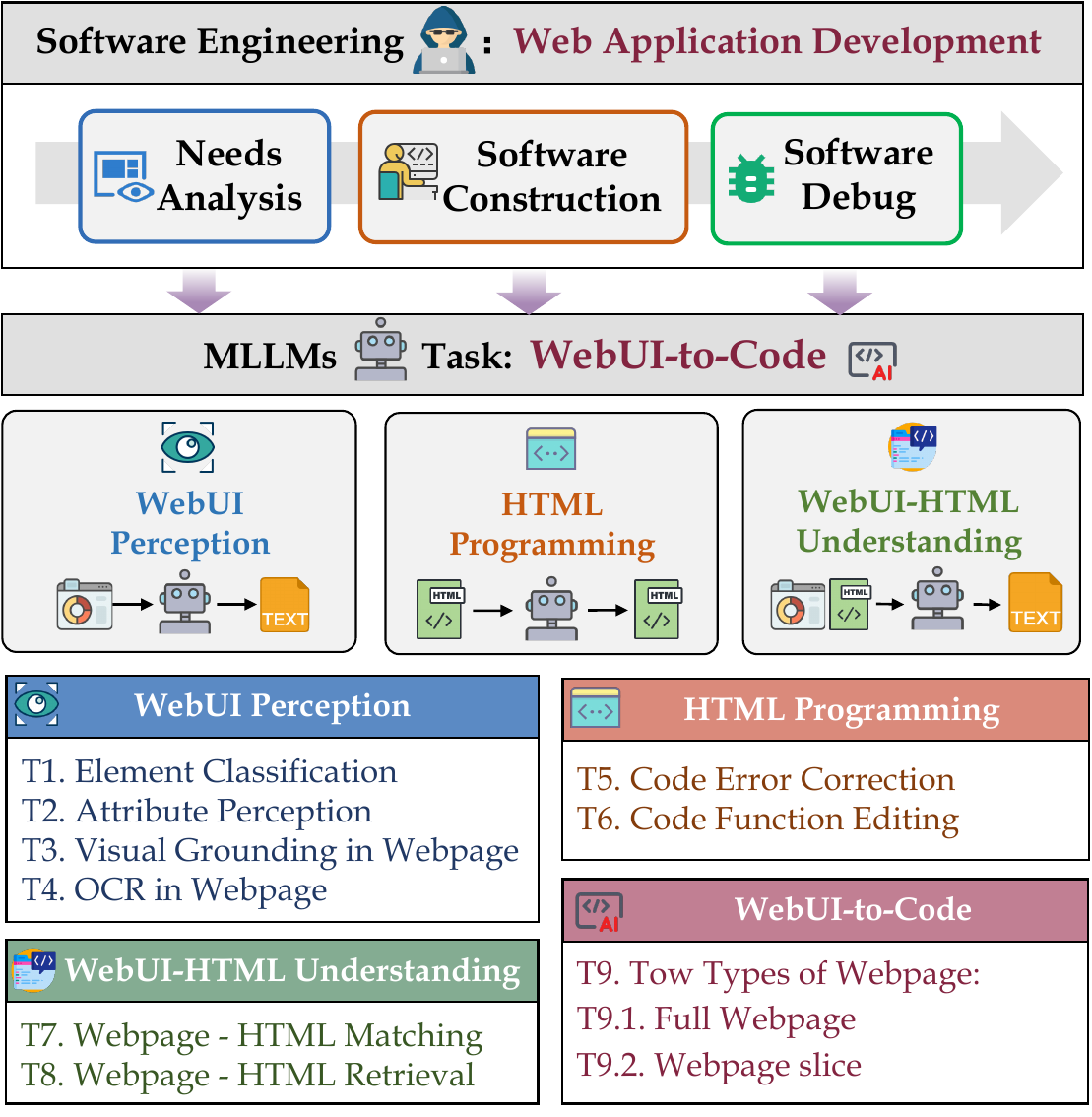}
    \caption{Evaluation taxonomy of WebUIBench.}
    \label{fig:fig1}
\end{figure}

\section{Introduction}

The emergence of Large Language Models(LLMs) has rapidly reshaped the landscape of software engineering. AI code generation \cite{chen2024survey,shin2021survey,dehaerne2022code} evolves from assisting developers to independently completing the entire development lifecycle (\textit{i.e.,} AI software engineer). Automatic website development is a challenging and widely discussed multimodal code generation scenario\cite{si2024design2code,yun2024web2code,beltramelli2018pix2code}: Multimodal Large Language Models(MLLMs) are required to generate front-end code projects based on user-provided WebUI images (WebUI-to-Code).  

Recent works\cite{si2024design2code,yun2024web2code,beltramelli2018pix2code} have evaluated MLLMs and reached a consensus that MLLMs struggle to generate complex websites, revealing a significant gap between solutions and practical applications. Therefore, it is essential to identify the challenges across various development stages and evaluate the corresponding sub-capabilities of models. However, current benchmarks\cite{guo2024iwbench,si2024design2code} typically focus on assessing the output quality of generated website (\textit{e.g.,} webpage elements and layout) and mostly lack evaluation for sub-capabilities. To address this issue, \cite{yun2024web2code} propose webpage understanding benchmark, but they are restricted to only one type of sub-capability and limited to the dataset quality annotated by LLMs. Inspired by the main activities of software engineering\cite{biolchini2005systematic}, we initially propose a taxonomy for the capability evaluation of sub-capability as depicted in Figure \ref{fig:fig1}.



\begin{figure*}
    \centering
    \includegraphics[width=1.0\linewidth]{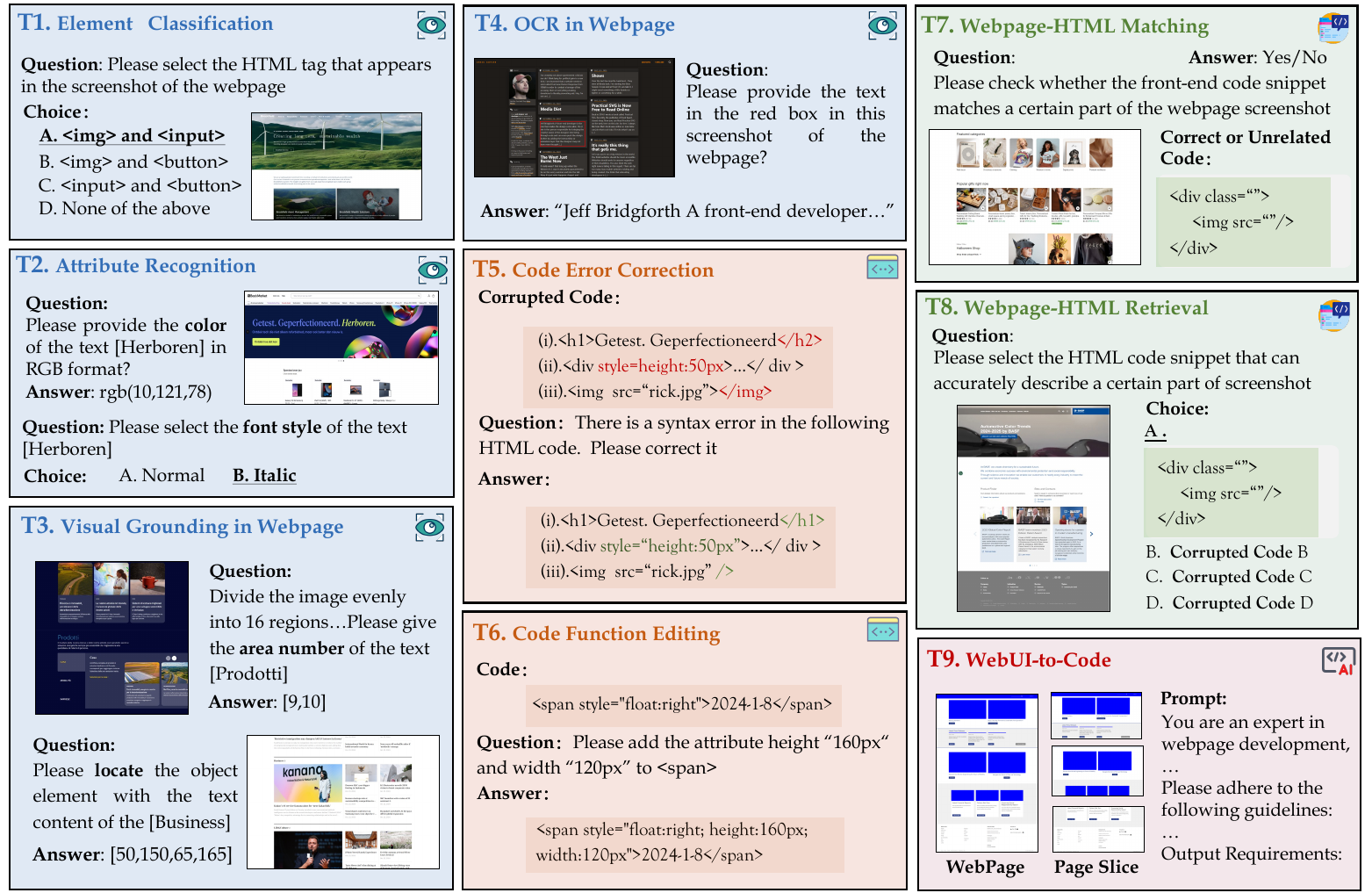}
    \caption{Task examples in the WebUI benchmark, from the \textbf{\textcolor[RGB]{50,107,182}{WebUI Perception}}, \textbf{\textcolor[RGB]{197,90,17}{HTML Programming}}, and \textbf{\textcolor[RGB]{84,130,53}{WebUI-HTML Understanding}} and \textbf{\textcolor[RGB]{167,90,121}{WebUI-to-Code}} task.}
    \label{fig:fig2}
\end{figure*}


Our core idea is to align the evaluation criteria of models’ capabilities with the task requirements of software engineering. To this end, we propose the following sub-capability evaluation dimensions: 
\noindent\textbf{(i) WebUI Perception:} WebUI images serve as visual carriers of development requirements. The fundamental skill of the model is to accurately perceive the visual semantic information in the webpage, including both text and images. \textbf{(ii) HTML Programming:} During the software construction phase, the model's knowledge reservoir and programming skills in front-end code are essential for assisting or substituting developers for efficient development. and \textbf{(iii) WebUI-HTML Understanding:} Post-software development, code testing and adjustments are necessary to ensure requirement accuracy. This necessitates the model's ability to perform cross-modality reasoning between design images and code functionalities. Furthermore, we draw from current leading MLLMs evaluation benchmarks\cite{liu2025mmbench,li2023seed,yue2024mmmu,li2024survey} to design multiple sub-tasks for each evaluation dimension, tailored to the characteristics of web data. In summary, our main contributions are three-fold:

\begin{itemize}
\item Construction of WebUIBench Dataset: The raw data is collected from 5 categories of frequently used real-world websites, including 719 complete webpage screenshots, source code and fine-grained information of all page elements. Based on this, WebUIBench consists of 2,488 webpage slices and 21,793 question-answer pairs across 9 sub-tasks.

\item Evaluation of Mainstream MLLMs: The evaluation process is conducted in 29 mainstream MLLMs, including 22 open-source models such as the InternVL2.5 series and the Qwen2-VL series with parameters ranging from 2B to 78B, and 7 closed-source MLLMs, such as GPT-4o, Gemini-1.5 Pro, and Claude-3.5-Sonnet.

\item Analysis of Challenges:  The primary conclusion is that most MLLMs are not capable of performing the complete front-end software development process as effectively as humans. The observed positive correlation between sub-capabilities and WebUI-to-Code performance validates our evaluation approach. It also reveals that the primary challenge for current MLLMs is to enhance and balance sub-capabilities across different dimensions.
\end{itemize}

\section{Taxonomy of Evaluation}
\label{sec:tax}
Inspired by \textbf{\textcolor[RGB]{167,90,121}{Web Application Development}}, our benchmark evaluates \textit{WebUI-to-Code(\textbf{Task9})} capability, and three essential sub-capabilities: \textit{WebUI Perception}, \textit{HTML Programming}, and \textit{WebUI-HTML Understanding}. For WebUI-to-Code, we provide two types of webpage: full webpage and webpage slice. For sub-capability evaluation, we designed various sub-tasks as follows:

\subsection{WebUI Perception}


\fcolorbox{white}{gray!15}{\parbox{.97\linewidth}{\textit{Inspiration:} The WebUI design is a visual representation of \textbf{\textcolor[RGB]{50,107,182}{Needs Analysis}}. WebUI Perception helps developers accurately grasp the requirements.}}

\vspace{2mm}


\noindent \textbf{\textcolor[RGB]{0,0,0}{Task1.} Element Classification.}  This task evaluates the model's capability to identify elements within webpage screenshots. The model must ascertain the presence of specific element types or combinations by fully understanding the screenshot.

\vspace{1mm}

\noindent \textbf{\textcolor[RGB]{0,0,0}{Task2.} Attribute Recognition.}  This task assesses the model's ability to discern detailed visual attributes of webpage elements, including text and background colors, font styles, and border styles.

\vspace{1mm}

\noindent \textbf{\textcolor[RGB]{0,0,0}{Task3.} Visual Grounding in Webpage.}  This task assess the model's capability to spatially locate elements on a webpage. We developed two levels of granularity for visual grounding tasks: \textit{(i)} At a coarse granularity, after evenly dividing the web page into a grid, the model identifies the grid region number of the specified element; \textit{(ii)} At a fine granularity, the model accurately returns the coordinates of the element's bounding box.

\vspace{2mm}

\noindent \textbf{\textcolor[RGB]{0,0,0}{Task4.} OCR in Webpage.}  This task tests the model's proficiency in extracting text from webpage screenshots. The model is required to detect and extract text content from a designated area framed by a red bounding box.

\subsection{HTML Programming}

\fcolorbox{white}{gray!15}{\parbox{.97\linewidth}{\textit{Inspiration:} The coding skills(e.g., HTML Programming) of developers ensure efficiency and stability throughout the \textbf{\textcolor[RGB]{197,90,17}{Software Construction}} lifecycle.}}

\vspace{2mm}

\noindent \textbf{\textcolor[RGB]{0,0,0}{Task5.} Code Error Correction.}  This task assesses the model's ability to correct syntax errors in front-end code. The model needs to identify errors in code snippets and return corrected versions.

\vspace{1mm}

\noindent \textbf{\textcolor[RGB]{0,0,0}{Task6.} Code Function Editing.}  This task evaluates the model's capability to implement static webpage functionalities through code. The model must edit and adjust code snippets according to the provided natural language instructions.

\subsection{WebUI-HTML Understanding}
\fcolorbox{white}{gray!15}{\parbox{.97\linewidth}{\textit{Inspiration:} \textbf{\textcolor[RGB]{84,130,53}{Software Debugging}} aims at ensuring consistency between the HTML code and the WebUI, reducing functional deficiencies through cross-modality understanding.}}

\vspace{2mm}

\noindent \textbf{Task7. Webpage-HTML Matching.}  The model determines whether the provided webpage screenshot and code snippet are correctly matched.

\vspace{1mm}

\noindent \textbf{Task8. Webpage-HTML Retrieval.}  The model selects the appropriate code snippet that corresponds to the given webpage screenshot from a selection of multiple snippets.

\vspace{1mm}

WebUI-HTML Understanding tasks directly simulate a critical prerequisite in debugging: developers must identify and locate relevant code sections based on visual interface elements before fixing bugs. MLLMs need cross-modally understand the matching relationship between WebUI images and HTML code ("MLLMs cannot directly use F12").

\section{Dataset}

\subsection{Raw Data Collection}
\label{sec:3-1}
WebUIBench consists of 5 categories of websites commonly visited by users: enterprise portals, background management systems, personal blogs, news sites, and e-commerce platforms. Firstly, we gather 1K websites (0.2K websites for each category) from the Internet. By using browser extension tools and manual collection, we collect the source HTML code and screenshot of these websites. Additionally, we extract detailed information of webpage elements, including tag categories, text content, CSS and spatial locations.

\vspace{1mm}

\noindent \textbf{Quality Control.} The incompletely or incorrectly loaded website are firstly reviewed and removed by human annotators
. For excessively long pages, often found in news sites and e-commerce platforms categories due to repetitive elements, we develop a page simplification algorithm to refactor source HTML code. The algorithm can streamline webpage elements and shorten page length while ensuring the quality and diversity of elements. Detailed information about the algorithm is provided in the Appendix \ref{apend:sec2}.

\subsection{Question and Answer Pairs Collection}
Users usually browse websites by scrolling up and down, similar to viewing through a "sliding window." Inspired by this observation, webpage slice is designed as the fundamental image data for constructing dataset. We segment the screenshot of webpage into slices of varying sizes based on the page layout and browser window size, ensuring each slice is relatively independent and semantically complete. The detailed segmentation algorithm is introduced  in the Appendix \ref{apend:sec2}. While collecting sliced screenshots, we also save the webpage element information within the slices and capture the corresponding code snippets to support the next step of the annotation process.

\vspace{1mm}

\begin{table}[t]
\small
\centering
\begin{tabular}{lr}
\toprule
\textbf{Statistic} & \textbf{Number} \\ \midrule
Total Website - HTML Code Samples & 719 \\
Total Question - Answer Samples  & 21793 \\
Total Website Types & 5 \\
Total Task Types   & 9 \\
\midrule
\multicolumn{2}{l}{\textit{Webpage Screenshots}} \\
\quad$\diamond$ Full page & 719 \\
\quad$\diamond$ Slice page & 2488 \\
\multicolumn{2}{l}{\textit{Screenshot Resolution}} \\
\quad$\diamond$ Maximal & 1800$\times$6802 \\
\quad$\diamond$ Minimal & 1800$\times$386 \\
\quad$\diamond$ Average & $\approx$1800$\times$1235 \\ \midrule
\multicolumn{2}{l}{\textit{Tokens of Question Captions}} \\
\quad$\diamond$ Maximal & 3582 \\
\quad$\diamond$ Minimal & 42 \\
\quad$\diamond$ Average & $\approx$287 \\
\midrule
Average slices per site & 3.46 \\
Average QA samples per slice & 10.68 \\
\bottomrule
\end{tabular}
\caption{Key statistics of WebUIBench.}
\label{tab:tab1}
\end{table}


\noindent \textbf{Automatic Labeling.} We first design QA templates for each task mentioned in Section \ref{sec:tax}, including question caption, options and answers. To enhance the diversity and challenge of the evaluation data, the QA templates for each task can be transformed into various forms based on the designed strategies (as shown in Appendix \ref{apend:sec3}). By retrieving element information and code snippets, QA templates are automatically filled as the complete evaluation samples. To ensure the standards and quality of the dataset, automatic labeling is conducted in multiple batches. Each task within a batch undergoes sampling inspection, and the generation process will be optimized until all sampled data pass the inspection.

\begin{figure}[t]
    \centering
    \includegraphics[width=0.99\linewidth]{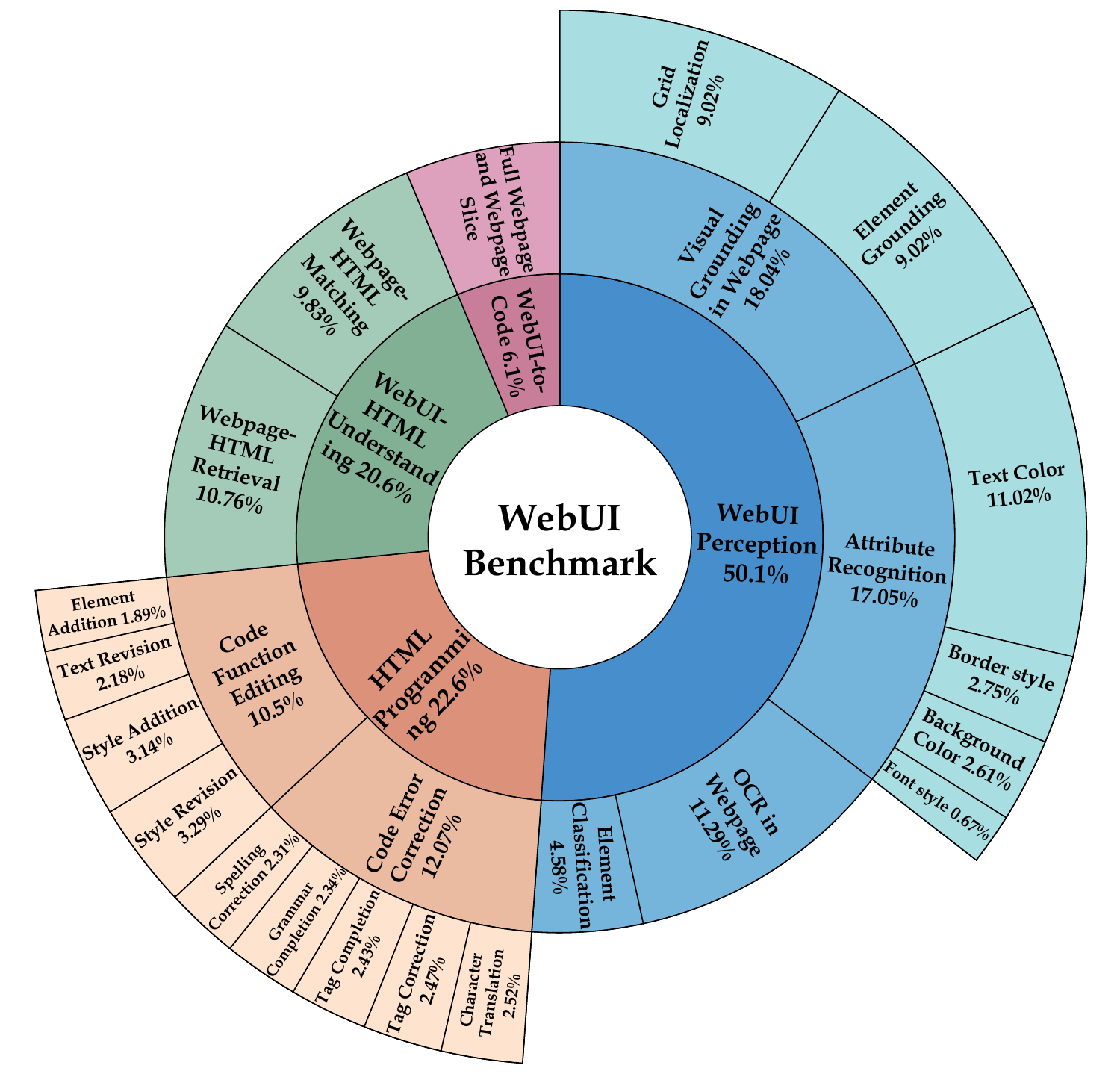}
    \caption{Question-Answer distribution of WebUIBench}
    \label{fig:fig3}
\end{figure}

\subsection{Dataset Statistics}

For webpage data collection, our dataset consists of 719 full webpage and 2488 webpage slices from 5 categories, covering a variety of resolution modes. We open-source the screenshot (\textit{.png} files), source HTML code (\textit{.html} files), and element information (\textit{.json} files) for these webpage. Based on this, WebUIBench includes 21,793 question-answer pairs, with an average of 10.68 question-answer pairs per webpage screenshot. Table \ref{tab:tab1} shows key statistics of dataset and Figure \ref{fig:fig3} shows the question-answer pairs distribution across different evaluation dimensions and tasks.




\vspace{1mm}

\begin{figure*}[t]
    \centering
    \includegraphics[width=0.98\linewidth]{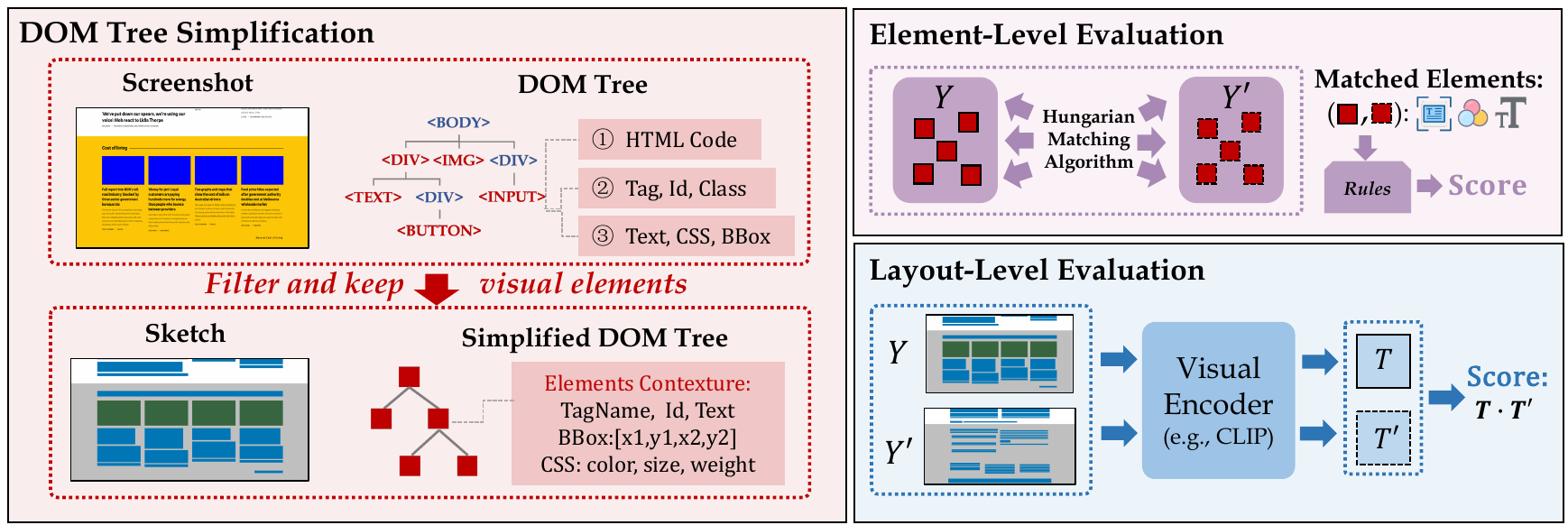}
    \caption{Schematic diagram of fine-grained WebUI-to-Code task evaluation process.}
    \label{fig:task10-eval}
\end{figure*}

\section{Metric}

\subsection{Automatic-metric Designs}

\textbf{Objective Question Scoring.} For multiple-choice tasks, we employ accuracy as the scoring metric. For open-ended tasks such as OCR, the score is given based on the text similarity between the generated string and the ground-truth string(character-level Sørensen-Dice similarity). It is also important to note that in code correction and code editing tasks, the clarity of the questions and prompts ensures unambiguous answers. Therefore, we also score these tasks by calculating the string similarity between the generated HTML code and the standard answer. 
Additionally, for tasks involving the identification of element color attributes, we use the CIEDE2000 color difference formula for scoring, following \cite{luo2001development}.

\vspace{2mm}

\noindent \textbf{WebUI-to-Code Task Scoring.} Evaluation is approached from two levels of granularity:

\vspace{1mm}
\noindent \textit{Coarse-Grained Evaluation:} This method involves calculating the visual similarity between the original webpage screenshot and the generated webpage screenshot to assess the overall visual quality. We utilize the visual pre-training backbone(e.g., CLIP\cite{radford2021learning}) to extract feature vectors and compute the cosine similarity as a measure of visual similarity.

\vspace{1mm}
\noindent \textit{Fine-Grained Evaluation:} We separate the fine-grained evaluation into element-level and layout-level assessments as shown in Figure \ref{fig:task10-eval}. The process includes: (i) Simplifying and restructuring the DOM tree of the webpage to preserve visual elements; (ii) Conducting evaluations separately at both the element and layout levels. The specific evaluation details are as follows:

\begin{itemize}
    \vspace{-1mm}
    \item \textit{DOM Tree Simplification:} We parse the original DOM tree to extract all elements related to visual presentation, including images, text, input fields, buttons, and areas with background colors. This results in a filtered set of webpage elements \( R = \{r_1, r_2, \ldots, r_m\} \). Based on the data preparation outlined in Section \ref{sec:3-1}, we gather corresponding information for each element in the set.
  \vspace{-1mm}
    \item \textit{Element-Level Evaluation:} Given the sets of elements for the real and generated webpages, \( R = \{r_1, r_2, \ldots, r_m\} \) and \( G = \{g_1, g_2, \ldots, g_n\} \), we construct a cost matrix based on the similarity of the text content of the elements, as referenced in \cite{si2024design2code}. The Hungarian algorithm is then used to find the optimal matching of elements. We evaluate similarity metrics for successfully matched elements from the real and generated webpages, including\textbf{text content, font color and background color}.
  \vspace{-1mm}
    \item \textit{Layout-Level Evaluation:} To maximize the decoupling of element-level and layout-level evaluations, we first remove content and style attributes of elements from the original webpage screenshot, preserving only size and spatial location information. Servel color blocks are used to distinguish elements of various tag categories, resulting in a sketch image of the webpage, as shown in Figure \ref{fig:task10-eval}. We then use visual pre-training backbone to extract visual features from this sketch image, quantifying the webpage layout information. Finally, we evaluate the effectiveness of layout generation by calculating the cosine similarity between the visual features of the real and generated webpage layouts.

\end{itemize}

\begin{table}[t]
\footnotesize
\centering
\caption{We evaluate the correlation between the rankings produced by our evaluation framework and those given by human experts. In our ablation study, we conducted three groups of experiments: (i) removing element-level evaluation (w/o Element), (ii) removing layout-level evaluation (w/o Layout), and (iii) retaining only the visual features from CLIP (CLIP).}

\scalebox{1.0}{
\begin{tabular}{lcccc}
\toprule
 & \textbf{Ours} & w/o Element & w/o Layout & CLIP  \\ \midrule

Human-1 & \textbf{0.83} & 0.75 & 0.81 & 0.76 \\
Human-2 & \textbf{0.81} & 0.74 & 0.76  & 0.75 \\
Human-3 & \textbf{0.85} & 0.78 & 0.81  & 0.78 \\ \midrule
\textit{Average} & \textbf{0.83} & 0.76 & 0.77 & 0.77  \\  \bottomrule
\label{tab:human}
\end{tabular}}
\end{table}

\begin{table*}[t]
\footnotesize
\centering
\caption{Evalutaion results of different MLLMs on the WebUIBench testset.  Bold entries represent the best performance in each category and the underline entries represent the second-best performance. Task name: EC=Element Classification, AP=Attribute Perception, VG=Visual Grounding, CEC=Code Error Correcting, CFE=Code Function Editing, WHM=WebUI-HTML Matching, WHR=WebUI-HTML Retrieval, W2C=WebUI-to-Code.}

\scalebox{0.97}{
\begin{tabular}{lc|ccccc|ccc|ccc|c}
\toprule

Model & Size & EC & OCR & AP & VG & \textbf{\textit{\textcolor[RGB]{50,107,182}{Avg.}}} & CEC & CFE & \textbf{\textit{\textcolor[RGB]{197,90,17}{Avg.}}} & WHM & WHR & \textbf{\textit{\textcolor[RGB]{84,130,53}{Avg.}}} &  \textbf{\textit{W2C}} \\ \hline
\multicolumn{14}{l}{\cellcolor[HTML]{c0c0c0}\textit{\textbf{Closed Source Model}}} \\
GPT-4o & - & 83.3 & 79.1 & 79.8 & 44.4 & \textbf{57.3} & 91.8 & 90.4 & \underline{91.1} & 65.7 & 41.9 & 53.8 & \textbf{82.0}  \\
GPT-4o-mini & - & 42.4 & 72.9 & 70.8 & 38.9 & 45.0 & 92.0 & 90.7 & \textbf{91.4} & 50.1 & 46.4 & 48.2 & 74.9 \\
Claude-3.5-Sonnet & - & 78.9 & 77.3 & 80.7 & 42.9& \underline{55.9} & 88.6 & 86.9& 87.8 & 73.7 & 43.6& \textbf{58.7} & \underline{80.2} \\
Gemini-1.5-pro  & - & 63.9  & 76.8 & 70.3 & 26.1& 47.4 & 87.7 & 84.8& 86.2 & 65.0 & 47.0& 56.0 & 80.0  \\
Yi-Vision & - & 59.2 & 37.7 & 68.9 & 28.5 & 38.9 & 84.3 & 81.5 & 82.9 & 46.3 & 48.9 & 47.6 & 77.0 \\
GLM-4v & -  & 62.4 & 62.7 & 62.6 & 29.4 & 43.5 & 53.3 & 32.3 & 42.8 & 48.4 & 65.0 & \underline{56.7} & 72.0 \\ 
Step-1.5v-mini & - & 62.1 & 49.4 & 57.1 & 17.6 & 46.5 & 86.6 & 84.9 & 85.7 & 44.8 & 53.8 & 49.3 & 67.9 \\
\hline
\multicolumn{14}{l}{\cellcolor[HTML]{c0c0c0}\textit{\textbf{Open Source Model}}} \\
Qwen2-VL & 2B & 28.2 & 49.6 & 54.4 & 39.9& 34.4 & 16.2 & 15.6& 15.9 & 24.7 & 29.8& 27.2 & 62.1 \\
InternVL2 & 2B & 46.8 & 34.5 & 49.4 & 29.8& 32.1 & 16.9 & 16.2& 16.5 & 28.2 & 39.1& 33.6 & 55.9 \\
InternVL2.5 & 2B & 43.8 & 45.5 & 41.6 & 37.8& 33.7 & 61.9 & 56.5& 59.2 & 31.0 & 58.7& 44.8 & 55.9 \\
Ovis1.6-Llama3.2 & 3B & 64.8 & 45.4 & 51.4 & 38.6& \underline{40.0} & 35.6 & 50.8& 43.2 & 18.5 & 37.4& 27.9 & \textbf{65.9} \\
InternVL2 & 4B & 56.7 & 47.4 & 57.3 & 39.7& \textbf{40.2} & 83.2 & 78.4& \underline{80.8} & 43.6 & 60.3& \underline{51.9} & 63.4\\
InternVL2.5 & 4B & 56.1 & 50.6 & 55.8 & 36.3& 39.8 & 92.1 & 86.9& \textbf{89.5} & 71.7 & 58.6& \textbf{65.2} & \underline{63.6} \\ \midrule
Qwen2-VL & 7B & 78.3 & 76.1 & 67.3 & 16.1& \textbf{47.6}  & 41.9 & 69.9& 55.9 & 55.9 & 36.2& 46.1 & 65.8 \\
InternVL2 & 8B & 32.4 & 54.8 & 59.2 & 40.9& 37.4 & 75.6 & 72.4& 74.0 & 57.9 & 62.9& \underline{60.4} & \underline{70.7} \\
InternVL2.5 & 8B & 26.0 & 47.5 & 60.2 & 42.4& 35.3 & 83.8 & 84.1& \textbf{83.9} & 75.4 & 62.9& \textbf{69.1} & \textbf{71.9} \\
MiniCPM-V-2.6 & 8B & 49.9 & 54.7 & 54.5 & 23.3& 36.5 & 66.2 & 62.3& 64.2 & 21.6 & 34.7& 28.1 & 70.4 \\
Phi-3-vision & 8B & 62.9 & 16.4 & 57.3 & 40.5& 35.5 & 58.1 & 44.5& 51.3 & 25.7 & 36.7& 31.2 & 56.0 \\
Phi-3.5-vision & 8B & 12.2 & 3.8 & 53.7 & 29.4& 19.8 & 67.9 & 45.2& 56.6 & 25.4 & 35.4& 30.4 & 53.5 \\
Ovis1.6-Gemma2 & 9B & 57.5 & 51.0 & 70.4 & 21.1& \underline{40.0} & 65.3 & 85.9& \underline{75.6} & 42.2 & 45.3& 43.8 & 69.8\\ \midrule
InternVL2 & 26B & 31.1 & 65.0 & 65.4 & 35.3& 39.4 & 80.8 & 82.3& 81.6 & 52.7 & 55.0& 53.8 & 67.6\\
InternVL2.5 & 26B & 65.9 & 57.1 & 67.9 & 53.6& \textbf{49.1} & 91.5 & 89.9& \underline{90.7} & 80.3 & 64.2& \underline{72.2} & \textbf{75.3} \\
Ovsi1.6-Gemma2 & 27B & 42.8 & 42.9 & 46.1 & 18.9& 30.2 & 89.9 & 88.9& 89.4 & 52.7 & 42.5& 47.6 & 74.0 \\
InternVL2.5 & 38B & 36.1 & 55.9 & 49.5 & 40.5& 36.5 & 92.4 & 90.5& \textbf{91.5} & 89.7 & 65.9& \textbf{77.9} & \underline{74.6}\\
InternVL2 & 40B & 50.8 & 66.1 & 61.6 & 29.7& \underline{41.8} & 76.7 & 73.1& 74.9 & 65.7 & 60.9& 63.3 & 74.3\\ \midrule
Qwen2-VL & 72B & 71.1 & 83.4 & 78.9 & 24.5& \textbf{51.6} & 59.5 & 64.7& 62.1 & 74.7 & 49.8& \underline{62.2} & \textbf{77.3}\\
NVLM-D & 72B & 81.5 & 14.4 & 63.9 & 57.7& 43.5 & 30.7 & 25.7& 28.2 & 66.0 & 44.9& 55.5 & 62.3\\
InternVL2 & 76B & 42.3 & 71.3 & 66.0 & 39.3& 43.8 & 83.5 & 86.1& \underline{84.8} & 61.9 & 57.8& 59.9 & 74.9\\
InternVL2.5 & 78B & 41.6 & 62.0 & 73.8 & 43.5& \underline{44.2} & 92.5 & 88.9& \textbf{90.7} & 83.7 & 59.7& \textbf{71.7} & \underline{75.7}\\ \bottomrule
\label{tab:main_result}
\end{tabular}}
\end{table*}


\subsection{Human-centric Evaluations}

We conduct an experiment to validate our evaluation framework: we randomly select 100 samples from the WebUI-to-Code task, and invite three front-end experts to rank the webpages generated by five models (GPT-4o, Qwen2-VL-72B, InternVL2.5-26B, Internvl2.5-8B, Ovis-Gemma2-3B). We calculate the correlation between human rankings and our evaluation framework rankings. Additionally, we conduct ablation experiments on different evaluation dimensions of our framework. The correlation results are as follows:

As shown in the Table \ref{tab:human}, \textbf{the strong correlation (correlation>0.8) with human expert confirms the validity and effectiveness of our evaluation framework}. Ablation experiment results also confirmed that adding element and layout dimensions significantly improved the webpage similarity comparison.

\section{Experiments}

\subsection{Models}

We select both the latest and top-performing MLLMs for evaluation, including closed-source models: GPT-4o \cite{hurst2024gpt}, GPT-4o-mini, Gemini-1.5-Pro-002\cite{team2024gemini}, Claude-3.5-Sonnet\cite{TheC3}, GLM-4V\cite{glm2024chatglm}, Yi-Vision\cite{young2024yi} and Step-1.5v\cite{step1_5}; open-source models: InternVL2.5 series\cite{chen2024expanding}, InternVL2 series\cite{chen2024expanding}, Qwen2-VL series\cite{wang2024qwen2vlenhancingvisionlanguagemodels}, Ovis-Gemma2 series\cite{lu2024ovis}, Phi-Vision series\cite{abdin2024phi}, NVLM-D-72B\cite{nvlm2024} and MiniCPM-V-2.6\cite{yao2024minicpm}. For open-source models, all parameter sizes within the same series are included in the evaluation process, ranging from the smallest model at 2 billion parameters to the largest at 78 billion parameters.

\begin{figure*}
    \centering
    \includegraphics[width=1.0\linewidth]{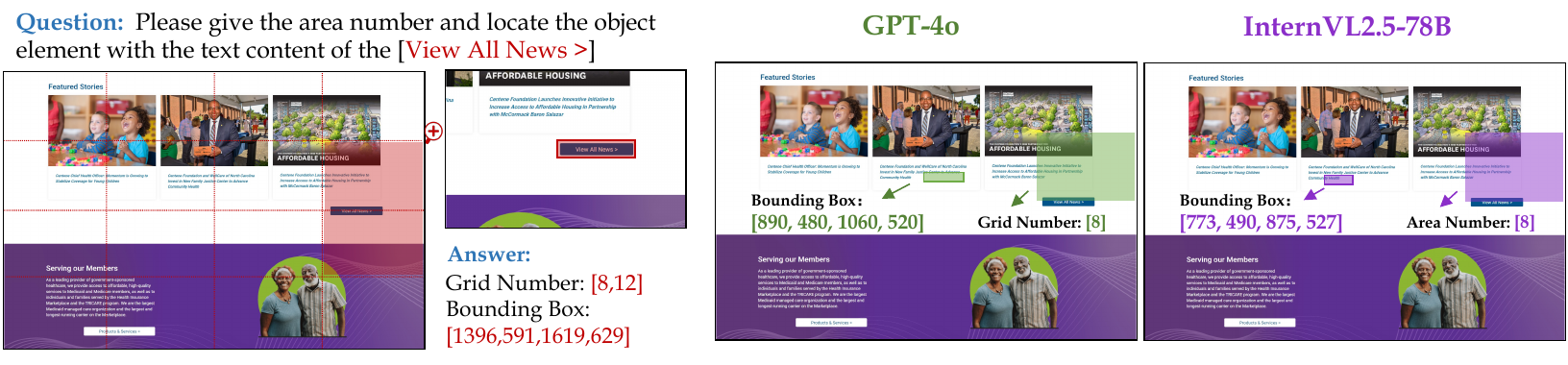}
    \caption{Visual grounding task examples (GPT-4o and InternVL2.5-78B).}
    \label{fig:vg-example}
\end{figure*}

\begin{figure*}
    \centering
    \begin{minipage}{0.48\textwidth}
        \centering
        \includegraphics[width=\linewidth]{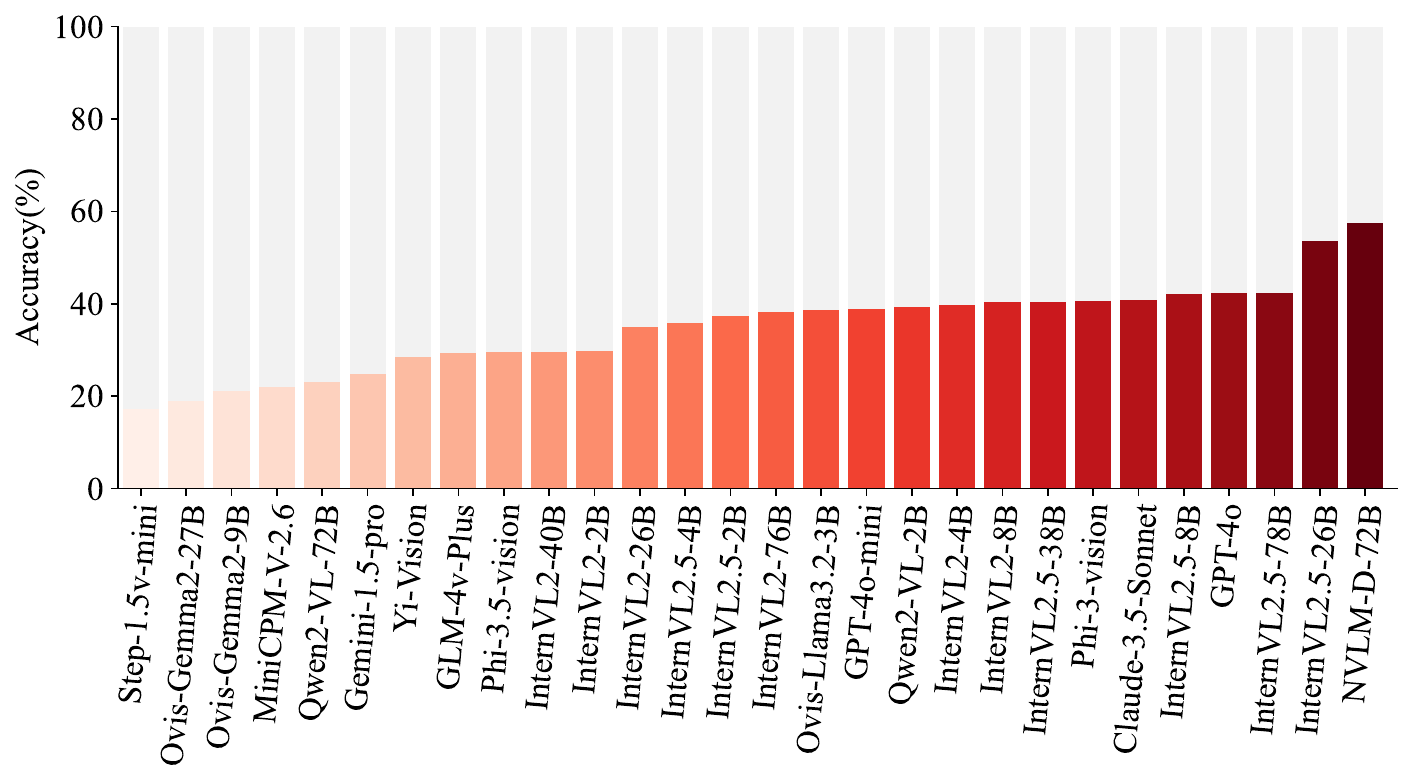}
        \caption{Results of grid number prediction (coarse-grained visual grounding)}
    \label{fig:vg1}
    \end{minipage}\hfill
    \begin{minipage}{0.48\textwidth}
        \centering
        \includegraphics[width=\linewidth]{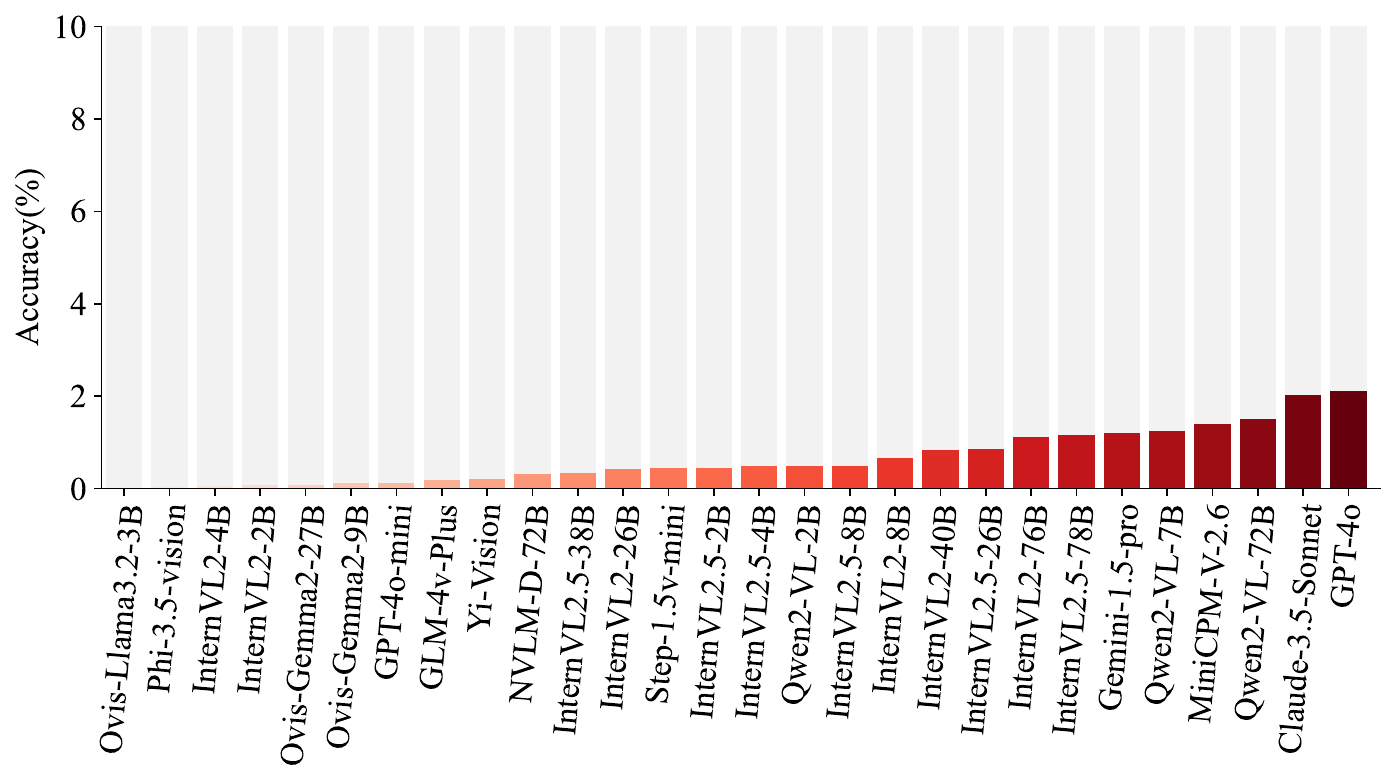}
        \caption{Results of bounding box prediction (fine-grained visual grounding)}
    \label{fig:vg2}
    \end{minipage}
\end{figure*}

\subsection{Main Results}

\vspace{-2mm}


As illustrated in Table \ref{tab:main_result}, we report the performance of all models across 9 tasks, with mean statistics calculated across various evaluation dimensions. To further provide a fine-grained evaluation and insights into the models' capabilities, we conduct a detailed analysis of both quantitative statistics and qualitative examples below.


\subsubsection{Analysis of Capability Characteristics }


The sub-capability assessment involves the results of \textit{WebUI Perception}, \textit{HTML Programming}, and \textit{WebUI-HTML Understanding} (\textit{i.e.,} from Task1 to Task8)

\vspace{2mm}

\noindent\textbf{MLLMs exhibit personalized development capability advantages. } For instance, \textit{Qwen2-VL} series models generally performs better on WebUI Perception dimensions, indicating proficiency in addressing challenges from the visual modality. Conversely,  \textit{InternVL2.5} series demonstrates a more pronounced advantage in HTML programming tasks. Overall, \textit{GPT-4o} exhibits a more comprehensive capability, yet remains weaker in WebUI-HTML Understanding tasks compared to the \textit{Claude-3.5-sonnet}. This phenomenon indicates that our proposed evaluation taxonomy can uncover finer-grained differences between models, which is beneficial for leveraging and enhancing personalized capabilities.

\vspace{2mm}



\noindent\textbf{Limitations of MLLMs in visual grounding task on Webpage. } A common weakness exhibited by most MLLMs is their difficulty in performing visual grounding tasks, with predicting bounding boxes being more challenging than predicting grid numbers. Figure \ref{fig:vg-example} shows a qualitative example where \textit{GPT-4o} and \textit{InternVL2.5-78B} can partially predict the area number of button locations correctly. However, the prediction of bounding boxes by both models completely deviated from the groundtruth. We attribute this phenomenon to the current MLLMs' lack of pixel-level understanding\cite{peng2024inst}, necessitating more fine-grained annotation and training with webpage images.



\vspace{1mm}

\noindent\textbf{MLLMs are poor at WebUI-HTML Understanding}. Recalling results in Table \ref{tab:main_result}, MLLMs (\textit{e.g.,} parameters$<$40B) perform a random guessing (\textit{i.e.,} score $\le$ 50\%) in Webpage-HTML Matching task, and performance in Webpage-HTML Retrieval tasks is also relatively lower compared to single-modality sub-tasks (\textit{e.g.,} HTML Programming task). We hypothesize that this capability deficiencies stems from the increased information density in cross-modality reasoning scenarios and
a lack of high-quality WebUI-HTML matching training data.

\begin{figure*}[ht]
    \centering
    \begin{minipage}{0.3\textwidth}
        \centering
        \includegraphics[width=\linewidth]{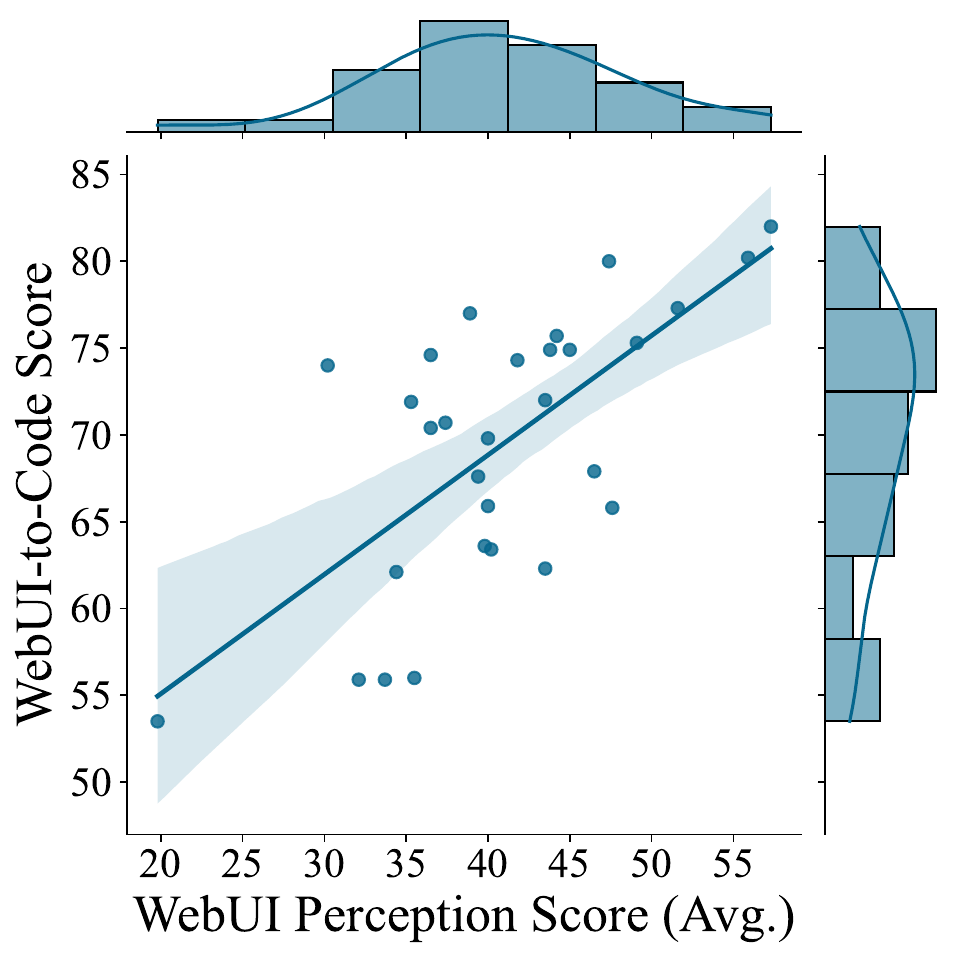}
    \end{minipage}\hfill
    \begin{minipage}{0.3\textwidth}
        \centering
        \includegraphics[width=\linewidth]{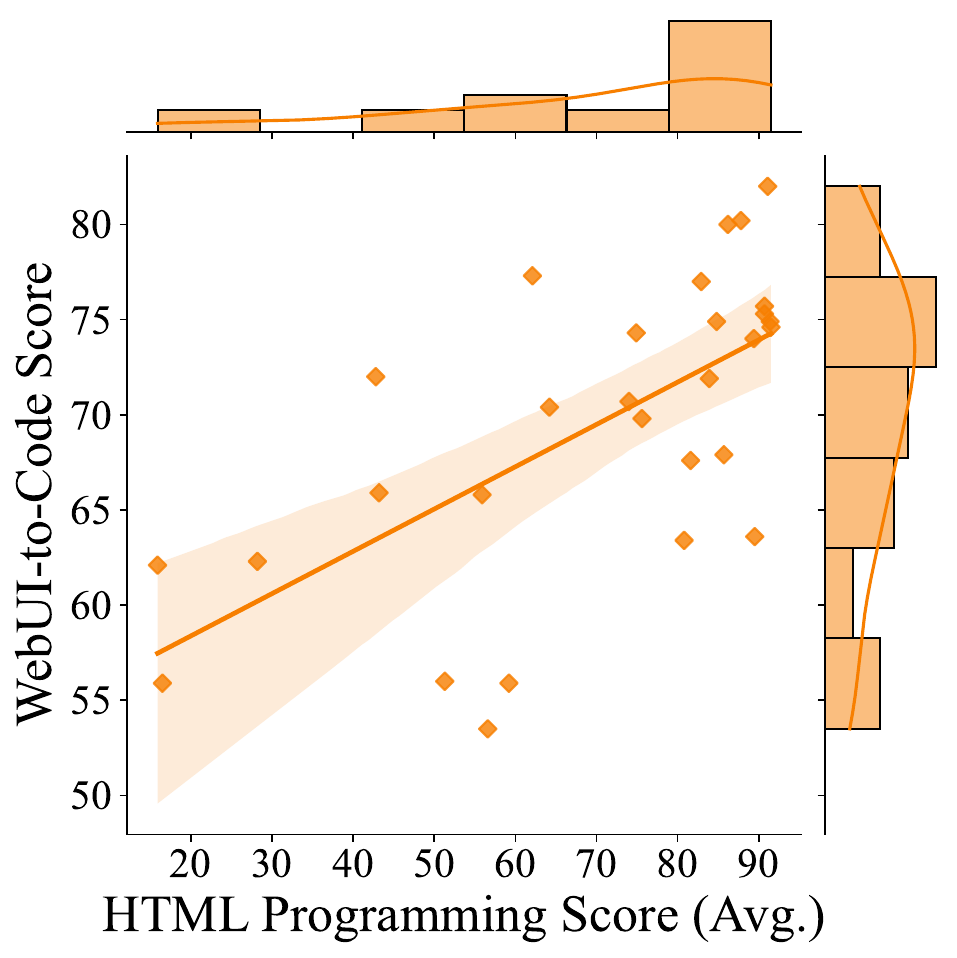}
    \end{minipage}\hfill
    \begin{minipage}{0.3\textwidth}
        \centering
        \includegraphics[width=\linewidth]{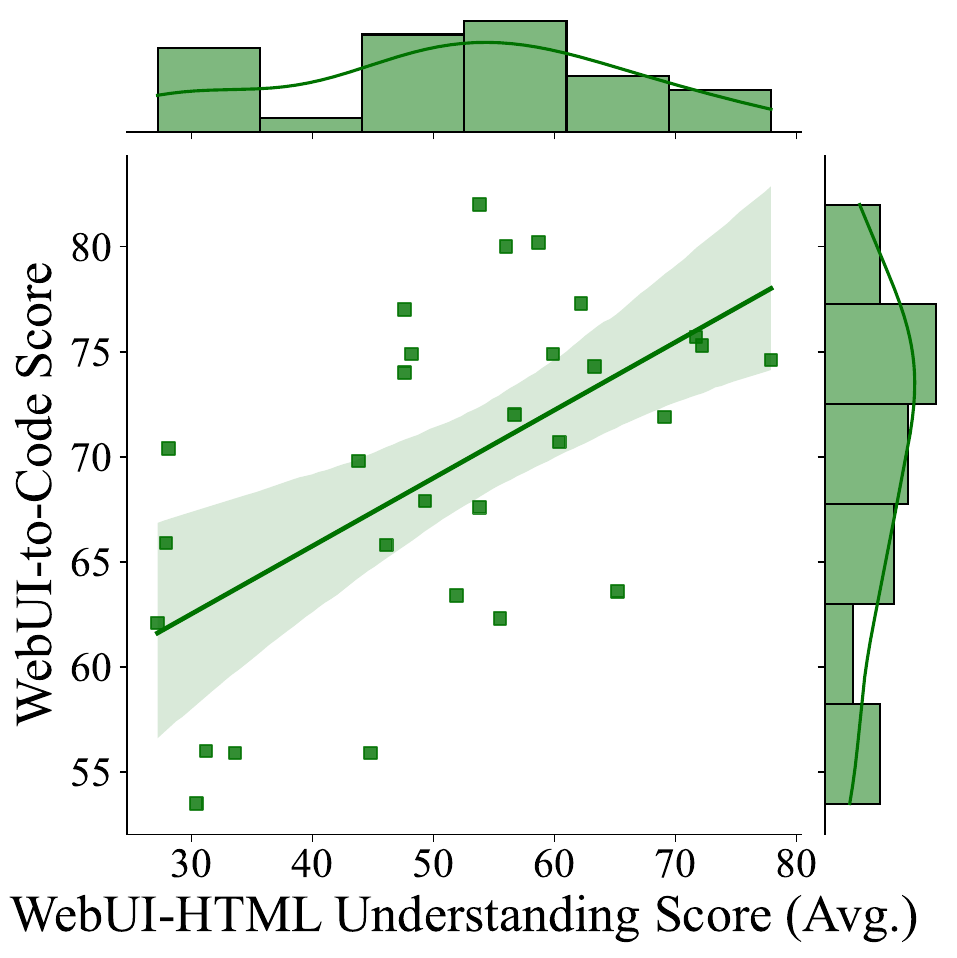}
    \end{minipage}\hfill
    \caption{Positive correlations between WebUI-to-Code performance and sub-capability performance.}
    \label{fig:pc}
\end{figure*}


\subsubsection{Results for WebUI-to-Code Task}

 The more results of element-level and layout-level evaluation are reported in Appendix.

\vspace{2mm}

\noindent\textbf{Positive correlations between WebUI-to-Code performance and personalized sub-capability.} As shown in Figure \ref{fig:pc}, the positive correlation indicates that our evaluation taxonomy effectively reveals the model's WebUI-to-Code capabilities across different sub-capability dimensions. We can initially use this phenomenon to analyze and explain the performance gap. For example, the low competitiveness of  \textit{NVLM-D-72B} among 70B+ models may be due to its deficiency in HTML programming capabilities (\textit{i.e,} CEC Score is 30.7\% and CFE Score is 25.7\%). It validates our idea of evaluating sub-capabilities according to software engineering principles.


\begin{table}[t]
\footnotesize
\centering
\caption{The results of the instruction-following failure rate and code compilation success rate for the small MLLMs.}

\scalebox{1.0}{
\begin{tabular}{lccc}
\toprule
Model & Size & $\sharp$Samples & Accuracy  \\ \midrule

\multicolumn{4}{l}{\textit{\textbf{Instruction-Following Evaluation ($\downarrow$)}}} \\
InternVL2 & 4B & 49 & 2.39\% \\
InternVL2.5 & 4B & 71 & 3.46\%  \\
InternVL2 & 2B & 811 & 3.96\%  \\
Qwen2-VL & 2B & 1,085 & 5.17\%  \\
InternVL2.5 & 2B & 1,393 & 6.81\%  \\
Ovis1.6-Llama3.2 & 3B & 2,895 & 14.14\%  \\ \midrule

\multicolumn{4}{l}{\textit{\textbf{HTML Code Compilation ($\uparrow$)}}} \\
Ovis1.6-Llama3.2 & 3B & 823  & 62.37\%  \\
InternVL2 & 4B & 504 & 38.23\% \\
InternVL2.5 & 4B & 247 & 18.74\% \\
InternVL2.5 & 2B & 239  & 18.13\%  \\
Qwen2-VL & 2B & 155 & 11.76\% \\
InternVL2 & 2B & 53 & 4.02\% \\ \bottomrule
\label{tab:ccs}
\end{tabular}}
\end{table}

\vspace{2mm}

\noindent\textbf{Small MLLMs face increasing inference cost in HTML generation.} HTML typically describes webpage content in long text form, posing challenges to the model's inference process. As shown in Table \ref{tab:ccs}, although the outputs from the smaller models generally passed the instruction-following tests, the code content within the output often failed to compile successfully. We notice that small models tend to output repetitive content or incomplete code, affecting the proper closure of HTML tags. It hinders their ability to perform generation tasks well, despite having decent performance in some sub-capabilities.

\vspace{2mm}

\begin{figure*}[t]
    \centering
    \includegraphics[width=0.94
    \linewidth]{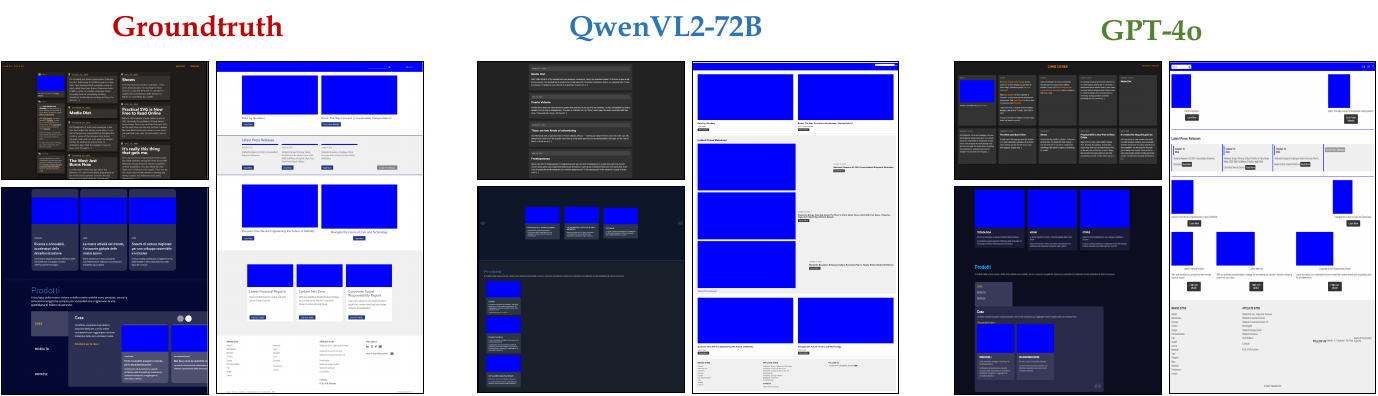}
    \caption{Examples of generated webpage by Qwen2-VL-72B and GPT4o on complex webpage and slices. Compared to implementing the styling of webpage elements, webpage layout remains a huge challenge for MLLMs.}
    \label{fig:example}
\end{figure*}

\noindent \textbf{Visualization for qualitative examples. } As shown in Figure \ref{fig:example}, we present the generation results of \textit{GPT-4o} and \textit{QwenVL2-72B} on complex webpages. By observing and comparing visual differences between generated webpage screenshots and WebUI images, we observe some interesting phenomena: (i)MLLMs demonstrates the ability to recognize vertical layouts of pages but struggles to identify and generate horizontal layouts. (ii) MLLMs can count elements effectively, yet performs poorly in generating the shapes and sizes of these elements. (iii) As the content of the webpage increases, these deficiencies become more pronounced.

\subsection{Explorations of Future Benchmarking}

\noindent \textbf{Front-end Frameworks Evaluation.} We select 100 WebUI images and used the Qwen2-VL-7B model to generate code, and setting whether to use the Tailwind framework (a lightweight front-end CSS framework) in the prompts. We present the compilation success rate and the webpage generation quality as follows:

Using the Tailwind framework improve the model's coding efficiency and success rate (47\% to 92\%), but unexpectedly, the quality of the generated webpages declined (0.718 to 0.697). Upon reviewing the code, we find syntax errors related to the this framework and instances where Tailwind was mixed with raw HTML inconsistently. This suggests that coding with Tailwind remains challenging for MLLMs. Additionally, deficiencies in HTML Programming also impact webpage generation quality, aligning with our findings on NLVM-D-72B.

\vspace{2mm}

\noindent \textbf{Dynamic Interaction Evaluation.} A comprehensive evaluation of JavaScript functionality is a major challenge. To explore the model’s capabilities in this area, we conducted a preliminary experiment to verify our hypothesis regarding their limitations: we designed 100 web development instructions requiring complex interactive functions, including form validation, event handling, and state management. These instructions covered various scenarios, from simple click events to complex data interactions. 

We then asked the leading models, Claude-3.7-Sonnet and GPT-4o, to generate the required code. Through manual interaction and testing, we observed that the models' success rate in completing these interactive functions was extremely low: Claude scored 0.26, while GPT-4o scored 0.20. These low scores confirm that current models significantly under-perform in generating functional JavaScript code.

\subsection{Discussions of Solutions}

In the front-end software development process, engineers usually construct the page layout first and then fill in the element information based on the completed layout. However, it seems that MLLMs internally couple the generation processes of both. Although the generation process is a black box, the similar rankings of MLLMs regarding element-level and layout-level scores may support this observations. Intuitively, element-level and layout-level information represent two types of patterns: local fine-grained features and global spatial features. Therefore, a reasonable hypothesis is that generating element and layout features simultaneously may not fully leverage the model's capabilities.

The above and discussions may suggest a future solution: decoupling webpage lay
out and element content generation into two steps, using methods like multimodal chain-of-thought\cite{wei2022chain} to incrementally generate webpages.

\section{Conclusion}
In this study, we introduce WebUIBench, a large-scale and comprehensive benchmark designed to evaluate the WebUI-to-Code capabilities of Multimodal Large Language Models (MLLMs). WebUIBench comprises over 21K question-answer pairs derived from more than 0.7K real-world websites, encompassing 9 distinct subtasks. We conducted extensive experiments on 7 state-of-the-art closed-source and 22 prominent open-source MLLMs. Our key findings highlight the models' deficiencies in webpage generation tasks across various dimensions, including cross-modality reasoning, element localization, and webpage layout generation. This benchmark provides critical insights and guidance for future research aimed at improving webpage generation performance.

\newpage
\section*{Limitations}
WebUIBench currently has the following limitations: (i) Imbalanced Data Distribution Across Sub-tasks: After manual and algorithmic filtering, some evaluation dimensions have only a few question-answer pairs remaining (\textit{e.g.,} the evaluation data for font-style accounts for only 0.67\%). In future work, we plan to address this by collecting new website data targeted at these missing categories. (ii) Lack of Mobile Webpage Evaluation Datasets: We have not yet constructed evaluation datasets for mobile platforms (\textit{e.g.,} smartphones). Considering that mobile web development is a prevalent task in software engineering, we plan to supplement our current data to include mobile evaluation datasets and results. (iii) Absence of Page Functionality Interaction Evaluation: In practical development, both static page and dynamic interaction functionalities need to be considered. In future work, it will also be important to evaluate MLLMs in generating interactive functionality code (\textit{e.g.,} JavaScript).

\bibliographystyle{plain} 
\bibliography{acl_latex.bib}

\begin{thebibliography}{32}
\providecommand{\natexlab}[1]{#1}

\bibitem[{ste()}]{step1_5}

\newblock \href {https://platform.stepfun.com/docs/llm/text} {step-1.5v-mini}.

\bibitem[{Abdin et~al.(2024)Abdin, Aneja, Awadalla, Awadallah, Awan, Bach, Bahree, Bakhtiari, Bao, Behl et~al.}]{abdin2024phi}
Marah Abdin, Jyoti Aneja, Hany Awadalla, Ahmed Awadallah, Ammar~Ahmad Awan, Nguyen Bach, Amit Bahree, Arash Bakhtiari, Jianmin Bao, Harkirat Behl, et~al. 2024.
\newblock Phi-3 technical report: A highly capable language model locally on your phone.
\newblock \emph{arXiv preprint arXiv:2404.14219}.

\bibitem[{Anthropic()}]{TheC3}
Anthropic.
\newblock \href {https://api.semanticscholar.org/CorpusID:268232499} {The claude 3 model family: Opus, sonnet, haiku}.

\bibitem[{Beltramelli(2018)}]{beltramelli2018pix2code}
Tony Beltramelli. 2018.
\newblock pix2code: Generating code from a graphical user interface screenshot.
\newblock In \emph{Proceedings of the ACM SIGCHI symposium on engineering interactive computing systems}, pages 1--6.

\bibitem[{Biolchini et~al.(2005)Biolchini, Mian, Natali, and Travassos}]{biolchini2005systematic}
Jorge Biolchini, Paula~Gomes Mian, Ana Candida~Cruz Natali, and Guilherme~Horta Travassos. 2005.
\newblock Systematic review in software engineering.
\newblock \emph{System engineering and computer science department COPPE/UFRJ, Technical Report ES}, 679(05):45.

\bibitem[{Chen et~al.(2024{\natexlab{a}})Chen, Guo, Jia, Zeng, Wang, Xu, Wu, Wang, Gao, Wang et~al.}]{chen2024survey}
Liguo Chen, Qi~Guo, Hongrui Jia, Zhengran Zeng, Xin Wang, Yijiang Xu, Jian Wu, Yidong Wang, Qing Gao, Jindong Wang, et~al. 2024{\natexlab{a}}.
\newblock A survey on evaluating large language models in code generation tasks.
\newblock \emph{arXiv preprint arXiv:2408.16498}.

\bibitem[{Chen et~al.(2024{\natexlab{b}})Chen, Wang, Cao, Liu, Gao, Cui, Zhu, Ye, Tian, Liu et~al.}]{chen2024expanding}
Zhe Chen, Weiyun Wang, Yue Cao, Yangzhou Liu, Zhangwei Gao, Erfei Cui, Jinguo Zhu, Shenglong Ye, Hao Tian, Zhaoyang Liu, et~al. 2024{\natexlab{b}}.
\newblock Expanding performance boundaries of open-source multimodal models with model, data, and test-time scaling.
\newblock \emph{arXiv preprint arXiv:2412.05271}.

\bibitem[{Dai et~al.(2024)Dai, Lee, Wang, Yang, Liu, Barker, Rintamaki, Shoeybi, Catanzaro, and Ping}]{nvlm2024}
Wenliang Dai, Nayeon Lee, Boxin Wang, Zhuolin Yang, Zihan Liu, Jon Barker, Tuomas Rintamaki, Mohammad Shoeybi, Bryan Catanzaro, and Wei Ping. 2024.
\newblock Nvlm: Open frontier-class multimodal llms.
\newblock \emph{arXiv preprint}.

\bibitem[{Dehaerne et~al.(2022)Dehaerne, Dey, Halder, De~Gendt, and Meert}]{dehaerne2022code}
Enrique Dehaerne, Bappaditya Dey, Sandip Halder, Stefan De~Gendt, and Wannes Meert. 2022.
\newblock Code generation using machine learning: A systematic review.
\newblock \emph{Ieee Access}, 10:82434--82455.

\bibitem[{GLM et~al.(2024)GLM, Zeng, Xu, Wang, Zhang, Yin, Zhang, Rojas, Feng, Zhao et~al.}]{glm2024chatglm}
Team GLM, Aohan Zeng, Bin Xu, Bowen Wang, Chenhui Zhang, Da~Yin, Dan Zhang, Diego Rojas, Guanyu Feng, Hanlin Zhao, et~al. 2024.
\newblock Chatglm: A family of large language models from glm-130b to glm-4 all tools.
\newblock \emph{arXiv preprint arXiv:2406.12793}.

\bibitem[{Guo et~al.(2024)Guo, Zhang, Chen, Gu, Yang, Du, Hui, Liu, Ma, Zhou et~al.}]{guo2024iwbench}
Hongcheng Guo, Wei Zhang, Junhao Chen, Yaonan Gu, Jian Yang, Junjia Du, Binyuan Hui, Tianyu Liu, Jianxin Ma, Chang Zhou, et~al. 2024.
\newblock Iw-bench: Evaluating large multimodal models for converting image-to-web.
\newblock \emph{arXiv preprint arXiv:2409.18980}.

\bibitem[{Hurst et~al.(2024)Hurst, Lerer, Goucher, Perelman, Ramesh, Clark, Ostrow, Welihinda, Hayes, Radford et~al.}]{hurst2024gpt}
Aaron Hurst, Adam Lerer, Adam~P Goucher, Adam Perelman, Aditya Ramesh, Aidan Clark, AJ~Ostrow, Akila Welihinda, Alan Hayes, Alec Radford, et~al. 2024.
\newblock Gpt-4o system card.
\newblock \emph{arXiv preprint arXiv:2410.21276}.

\bibitem[{Lauren{\c{c}}on et~al.(2024)Lauren{\c{c}}on, Tronchon, and Sanh}]{laurenccon2024unlocking}
Hugo Lauren{\c{c}}on, L{\'e}o Tronchon, and Victor Sanh. 2024.
\newblock Unlocking the conversion of web screenshots into html code with the websight dataset.
\newblock \emph{arXiv preprint arXiv:2403.09029}.

\bibitem[{Lee et~al.(2023)Lee, Joshi, Turc, Hu, Liu, Eisenschlos, Khandelwal, Shaw, Chang, and Toutanova}]{lee2023pix2struct}
Kenton Lee, Mandar Joshi, Iulia~Raluca Turc, Hexiang Hu, Fangyu Liu, Julian~Martin Eisenschlos, Urvashi Khandelwal, Peter Shaw, Ming-Wei Chang, and Kristina Toutanova. 2023.
\newblock Pix2struct: Screenshot parsing as pretraining for visual language understanding.
\newblock In \emph{International Conference on Machine Learning}, pages 18893--18912. PMLR.

\bibitem[{Li et~al.(2023)Li, Wang, Wang, Ge, Ge, and Shan}]{li2023seed}
Bohao Li, Rui Wang, Guangzhi Wang, Yuying Ge, Yixiao Ge, and Ying Shan. 2023.
\newblock Seed-bench: Benchmarking multimodal llms with generative comprehension.
\newblock \emph{arXiv preprint arXiv:2307.16125}.

\bibitem[{Li et~al.(2024)Li, Chen, Shi, Xiao, and Chen}]{li2024survey}
Lin Li, Guikun Chen, Hanrong Shi, Jun Xiao, and Long Chen. 2024.
\newblock A survey on multimodal benchmarks: In the era of large ai models.
\newblock \emph{arXiv preprint arXiv:2409.18142}.

\bibitem[{Liu et~al.(2025)Liu, Duan, Zhang, Li, Zhang, Zhao, Yuan, Wang, He, Liu et~al.}]{liu2025mmbench}
Yuan Liu, Haodong Duan, Yuanhan Zhang, Bo~Li, Songyang Zhang, Wangbo Zhao, Yike Yuan, Jiaqi Wang, Conghui He, Ziwei Liu, et~al. 2025.
\newblock Mmbench: Is your multi-modal model an all-around player?
\newblock In \emph{European conference on computer vision}, pages 216--233. Springer.

\bibitem[{Lu et~al.(2024)Lu, Li, Chen, Xu, Luo, Zhang, and Ye}]{lu2024ovis}
Shiyin Lu, Yang Li, Qing-Guo Chen, Zhao Xu, Weihua Luo, Kaifu Zhang, and Han-Jia Ye. 2024.
\newblock Ovis: Structural embedding alignment for multimodal large language model.
\newblock \emph{arXiv:2405.20797}.

\bibitem[{Luo et~al.(2001)Luo, Cui, and Rigg}]{luo2001development}
M~Ronnier Luo, Guihua Cui, and Bryan Rigg. 2001.
\newblock The development of the cie 2000 colour-difference formula: Ciede2000.
\newblock \emph{Color Research \& Application: Endorsed by Inter-Society Color Council, The Colour Group (Great Britain), Canadian Society for Color, Color Science Association of Japan, Dutch Society for the Study of Color, The Swedish Colour Centre Foundation, Colour Society of Australia, Centre Fran{\c{c}}ais de la Couleur}, 26(5):340--350.

\bibitem[{Nguyen and Csallner(2015)}]{nguyen2015reverse}
Tuan~Anh Nguyen and Christoph Csallner. 2015.
\newblock Reverse engineering mobile application user interfaces with remaui (t).
\newblock In \emph{2015 30th IEEE/ACM International Conference on Automated Software Engineering (ASE)}, pages 248--259. IEEE.

\bibitem[{Peng et~al.(2024)Peng, Meng, Chen, Xie, Liu, Gui, Xu, Qiu, Wu, and Jiang}]{peng2024inst}
Wujian Peng, Lingchen Meng, Yitong Chen, Yiweng Xie, Yang Liu, Tao Gui, Hang Xu, Xipeng Qiu, Zuxuan Wu, and Yu-Gang Jiang. 2024.
\newblock Inst-it: Boosting multimodal instance understanding via explicit visual prompt instruction tuning.
\newblock \emph{arXiv preprint arXiv:2412.03565}.

\bibitem[{Pix2code()}]{pix2code1705generating}
Tony~Beltramelli Pix2code.
\newblock Generating code from a graphical user interface screenshot [electronic resource].
\newblock \emph{arXiv preprint arXiv:1705.07962}.

\bibitem[{Radford et~al.(2021)Radford, Kim, Hallacy, Ramesh, Goh, Agarwal, Sastry, Askell, Mishkin, Clark et~al.}]{radford2021learning}
Alec Radford, Jong~Wook Kim, Chris Hallacy, Aditya Ramesh, Gabriel Goh, Sandhini Agarwal, Girish Sastry, Amanda Askell, Pamela Mishkin, Jack Clark, et~al. 2021.
\newblock Learning transferable visual models from natural language supervision.
\newblock In \emph{International conference on machine learning}, pages 8748--8763. PMLR.

\bibitem[{Shin and Nam(2021)}]{shin2021survey}
Jiho Shin and Jaechang Nam. 2021.
\newblock A survey of automatic code generation from natural language.
\newblock \emph{Journal of Information Processing Systems}, 17(3):537--555.

\bibitem[{Si et~al.(2024)Si, Zhang, Yang, Liu, and Yang}]{si2024design2code}
Chenglei Si, Yanzhe Zhang, Zhengyuan Yang, Ruibo Liu, and Diyi Yang. 2024.
\newblock Design2code: How far are we from automating front-end engineering?
\newblock \emph{arXiv preprint arXiv:2403.03163}.

\bibitem[{Team et~al.(2024)Team, Georgiev, Lei, Burnell, Bai, Gulati, Tanzer, Vincent, Pan, Wang et~al.}]{team2024gemini}
Gemini Team, Petko Georgiev, Ving~Ian Lei, Ryan Burnell, Libin Bai, Anmol Gulati, Garrett Tanzer, Damien Vincent, Zhufeng Pan, Shibo Wang, et~al. 2024.
\newblock Gemini 1.5: Unlocking multimodal understanding across millions of tokens of context.
\newblock \emph{arXiv preprint arXiv:2403.05530}.

\bibitem[{Wang et~al.(2024)Wang, Bai, Tan, Wang, Fan, Bai, Chen, Liu, Wang, Ge, Fan, Dang, Du, Ren, Men, Liu, Zhou, Zhou, and Lin}]{wang2024qwen2vlenhancingvisionlanguagemodels}
Peng Wang, Shuai Bai, Sinan Tan, Shijie Wang, Zhihao Fan, Jinze Bai, Keqin Chen, Xuejing Liu, Jialin Wang, Wenbin Ge, Yang Fan, Kai Dang, Mengfei Du, Xuancheng Ren, Rui Men, Dayiheng Liu, Chang Zhou, Jingren Zhou, and Junyang Lin. 2024.
\newblock \href {https://arxiv.org/abs/2409.12191} {Qwen2-vl: Enhancing vision-language model's perception of the world at any resolution}.
\newblock \emph{Preprint}, arXiv:2409.12191.

\bibitem[{Wei et~al.(2022)Wei, Wang, Schuurmans, Bosma, Xia, Chi, Le, Zhou et~al.}]{wei2022chain}
Jason Wei, Xuezhi Wang, Dale Schuurmans, Maarten Bosma, Fei Xia, Ed~Chi, Quoc~V Le, Denny Zhou, et~al. 2022.
\newblock Chain-of-thought prompting elicits reasoning in large language models.
\newblock \emph{Advances in neural information processing systems}, 35:24824--24837.

\bibitem[{Yao et~al.(2024)Yao, Yu, Zhang, Wang, Cui, Zhu, Cai, Li, Zhao, He et~al.}]{yao2024minicpm}
Yuan Yao, Tianyu Yu, Ao~Zhang, Chongyi Wang, Junbo Cui, Hongji Zhu, Tianchi Cai, Haoyu Li, Weilin Zhao, Zhihui He, et~al. 2024.
\newblock Minicpm-v: A gpt-4v level mllm on your phone.
\newblock \emph{arXiv preprint arXiv:2408.01800}.

\bibitem[{Young et~al.(2024)Young, Chen, Li, Huang, Zhang, Zhang, Li, Zhu, Chen, Chang et~al.}]{young2024yi}
Alex Young, Bei Chen, Chao Li, Chengen Huang, Ge~Zhang, Guanwei Zhang, Heng Li, Jiangcheng Zhu, Jianqun Chen, Jing Chang, et~al. 2024.
\newblock Yi: Open foundation models by 01. ai.
\newblock \emph{arXiv preprint arXiv:2403.04652}.

\bibitem[{Yue et~al.(2024)Yue, Ni, Zhang, Zheng, Liu, Zhang, Stevens, Jiang, Ren, Sun et~al.}]{yue2024mmmu}
Xiang Yue, Yuansheng Ni, Kai Zhang, Tianyu Zheng, Ruoqi Liu, Ge~Zhang, Samuel Stevens, Dongfu Jiang, Weiming Ren, Yuxuan Sun, et~al. 2024.
\newblock Mmmu: A massive multi-discipline multimodal understanding and reasoning benchmark for expert agi.
\newblock In \emph{Proceedings of the IEEE/CVF Conference on Computer Vision and Pattern Recognition}, pages 9556--9567.

\bibitem[{Yun et~al.(2024)Yun, Lin, Thushara, Bhat, Wang, Jiang, Deng, Wang, Tao, Li et~al.}]{yun2024web2code}
Sukmin Yun, Haokun Lin, Rusiru Thushara, Mohammad~Qazim Bhat, Yongxin Wang, Zutao Jiang, Mingkai Deng, Jinhong Wang, Tianhua Tao, Junbo Li, et~al. 2024.
\newblock Web2code: A large-scale webpage-to-code dataset and evaluation framework for multimodal llms.
\newblock \emph{arXiv preprint arXiv:2406.20098}.

\end{thebibliography}

\vspace{140mm}

\appendix

\section{Appendix}
\label{sec:appendix}

\subsection{Related Work}
Before the advent of Large Language Models(LLMs), a series of works\cite{beltramelli2018pix2code,lee2023pix2struct,nguyen2015reverse,pix2code1705generating} have already begun exploring how to convert webpage screenshots into HTML code. With the development of Multimodel Large Language Models(MLLMs), this field has seen the emergence of several works aimed at evaluating and addressing webpage code generation issues: WebSight\cite{laurenccon2024unlocking} introduced a large-scale synthetic dataset to train models in the code generation domain, but did not provide an evaluation dataset or methodology. Design2Code\cite{si2024design2code} was the first to systematically evaluate both open-source and closed-source MLLMs using real webpage data. Web2Code\cite{yun2024web2code} proposed an evaluation task for webpage understanding, expanding previous evaluation frameworks and introducing a high-quality code instruction dataset. IWBench\cite{guo2024iwbench} presented an evaluation method focused on webpage layout and improved generation algorithms using CoT. 


\begin{table}[h]
\footnotesize
\centering
\caption{Comparison of WebUIBench with previous works}

\scalebox{1.0}{
\begin{tabular}{lccc}
\toprule
Benchmark & Source & $\sharp$Size & Sub-cap.  \\ \midrule
Websight & Synthetic & 823K & \color{red}\ding{55} \\
Pixel2Code & Synthetic & 1.7K & \color{red}\ding{55}   \\
Web2Code & Synthetic & 884.7k & $\color{green}\checkmark$  \\ \midrule
Design2Code & Real-World & 484  & \color{red}\ding{55}  \\
IWBench & Real-World & 1.2K  & \color{red}\ding{55} \\
\textbf{WebUIBench(Ours)} & Real-World & 21K & $\color{green}\checkmark$ \\  \bottomrule
\label{tab:combench}
\end{tabular}}
\end{table}

\subsection{Examples for Different Tasks}
\begin{figure*}
    \centering
    \includegraphics[width=1.0\linewidth]{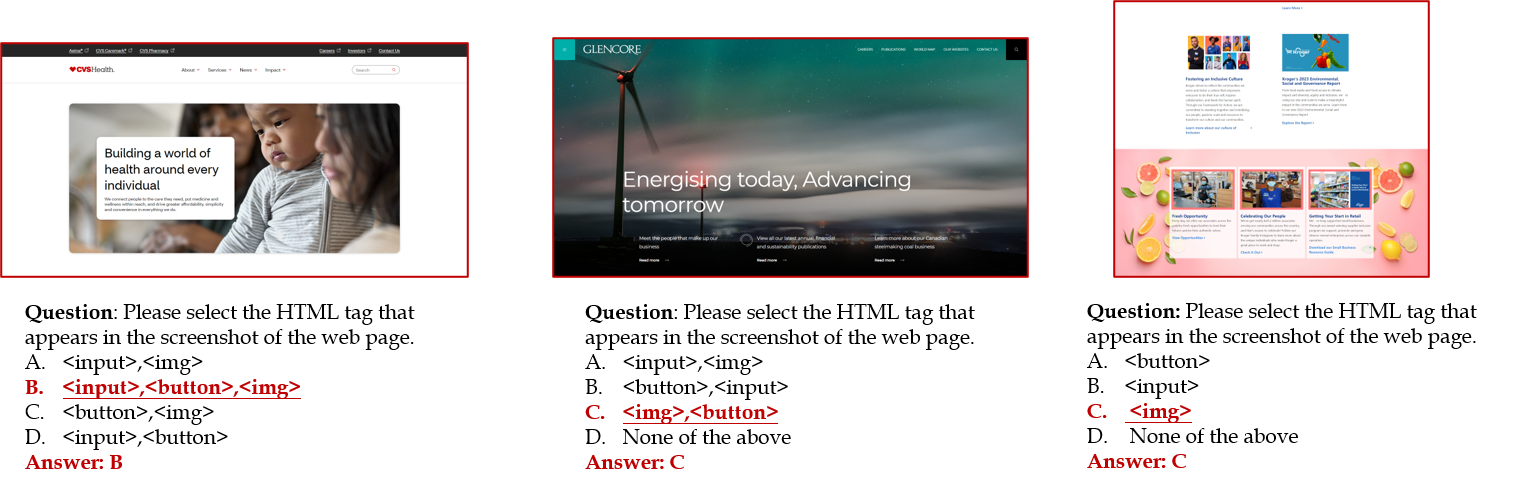}
    \caption{Samples of Element Classification.}
    \label{fig:task1}
\end{figure*}

As shown in Figure \ref{fig:task1} to \ref{fig:task7} , we provide specific examples for each evaluation task under the three major capability dimensions to illustrate the sample dataset.

\begin{figure*}
    \centering
    \includegraphics[width=1.0\linewidth]{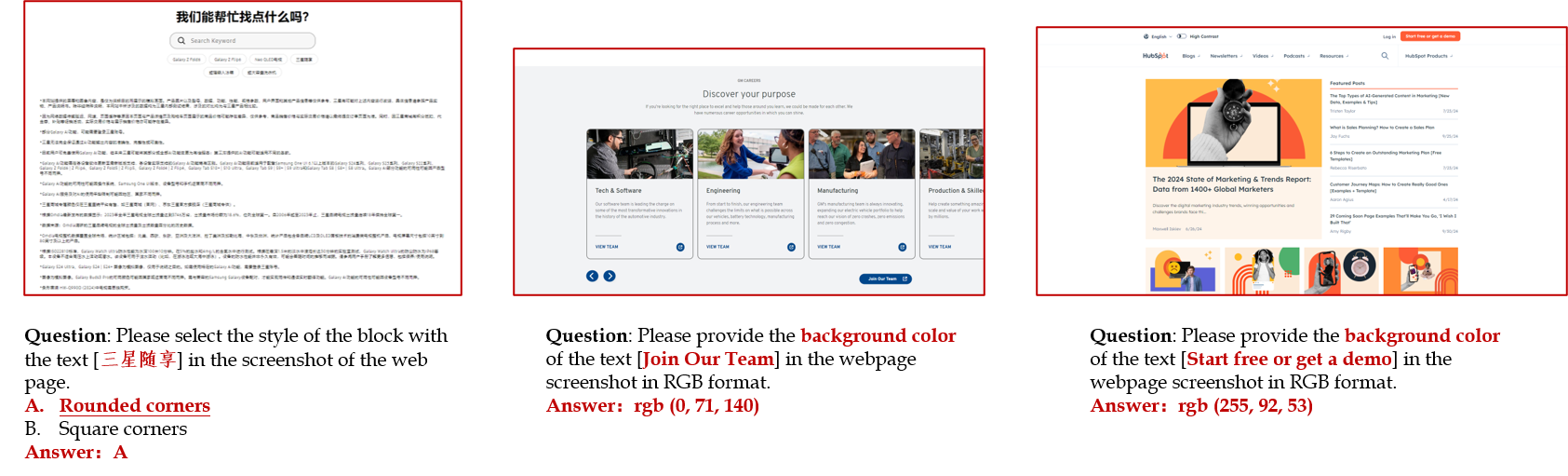}
    \caption{Samples of Attribute Perception.}
    \label{fig:vg-qa}
\end{figure*}

\begin{figure*}
    \centering
    \includegraphics[width=1.0\linewidth]{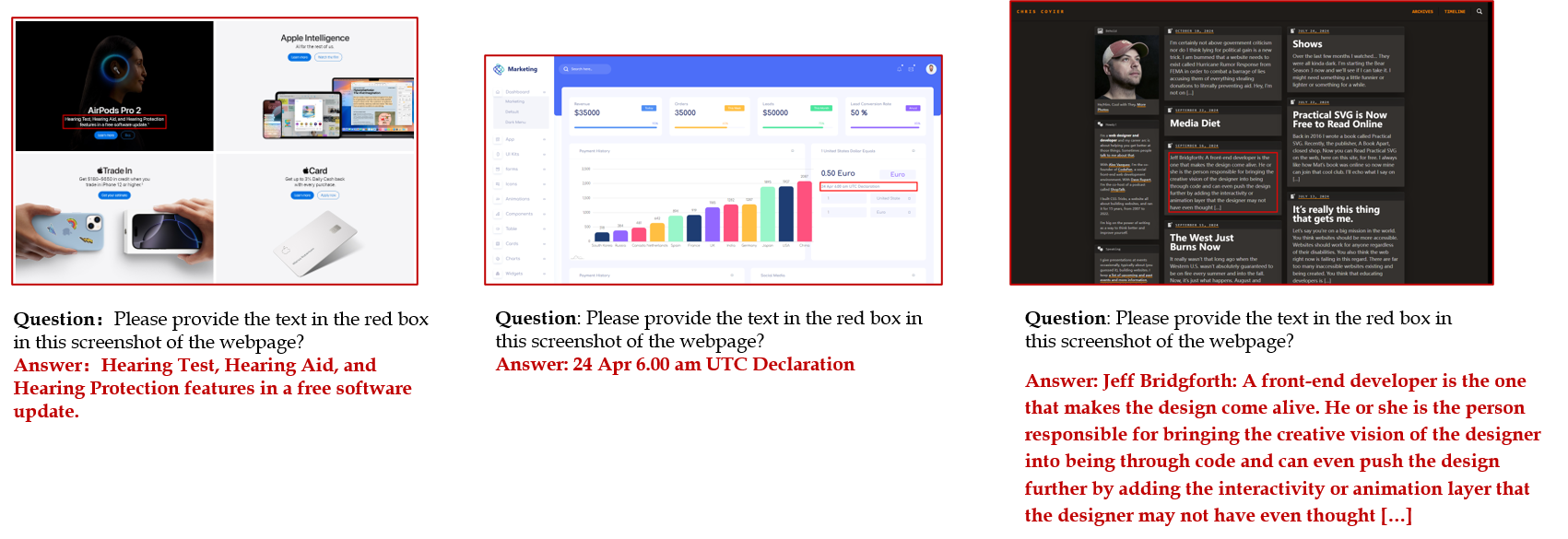}
    \caption{Samples of OCR in the Webpage.}
    \label{fig:vg-qa}
\end{figure*}

\begin{figure*}
    \centering
    \includegraphics[width=1.0\linewidth]{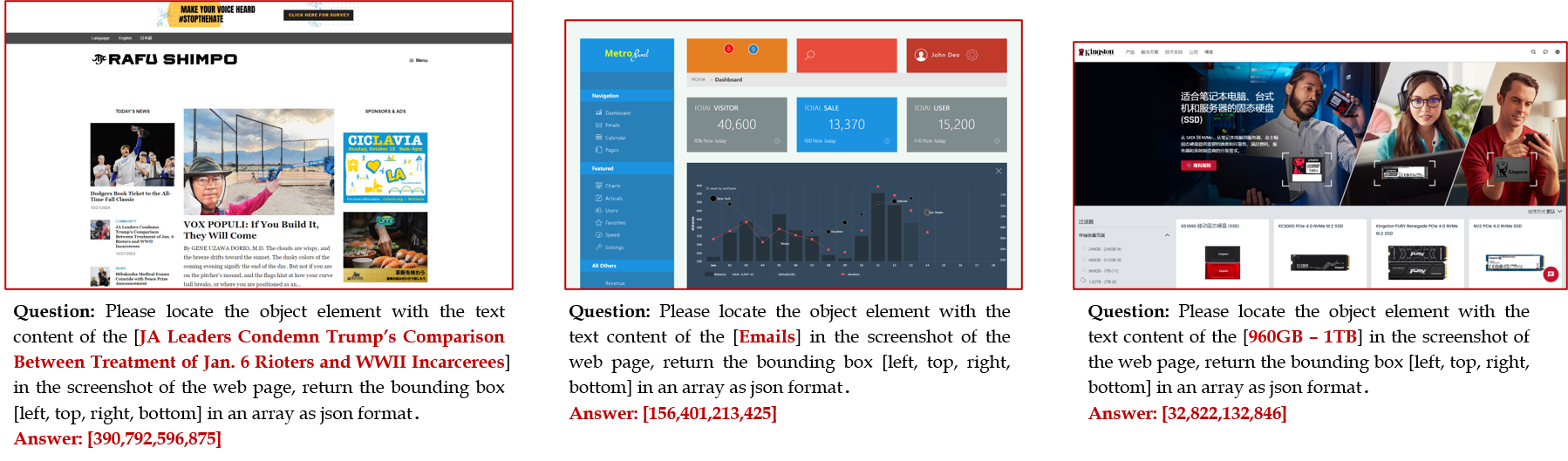}
    \caption{Samples of Visual Grounding(fine granularity).}
    \label{fig:vg-qa}
\end{figure*}

\begin{figure*}
    \centering
    \includegraphics[width=1.0\linewidth]{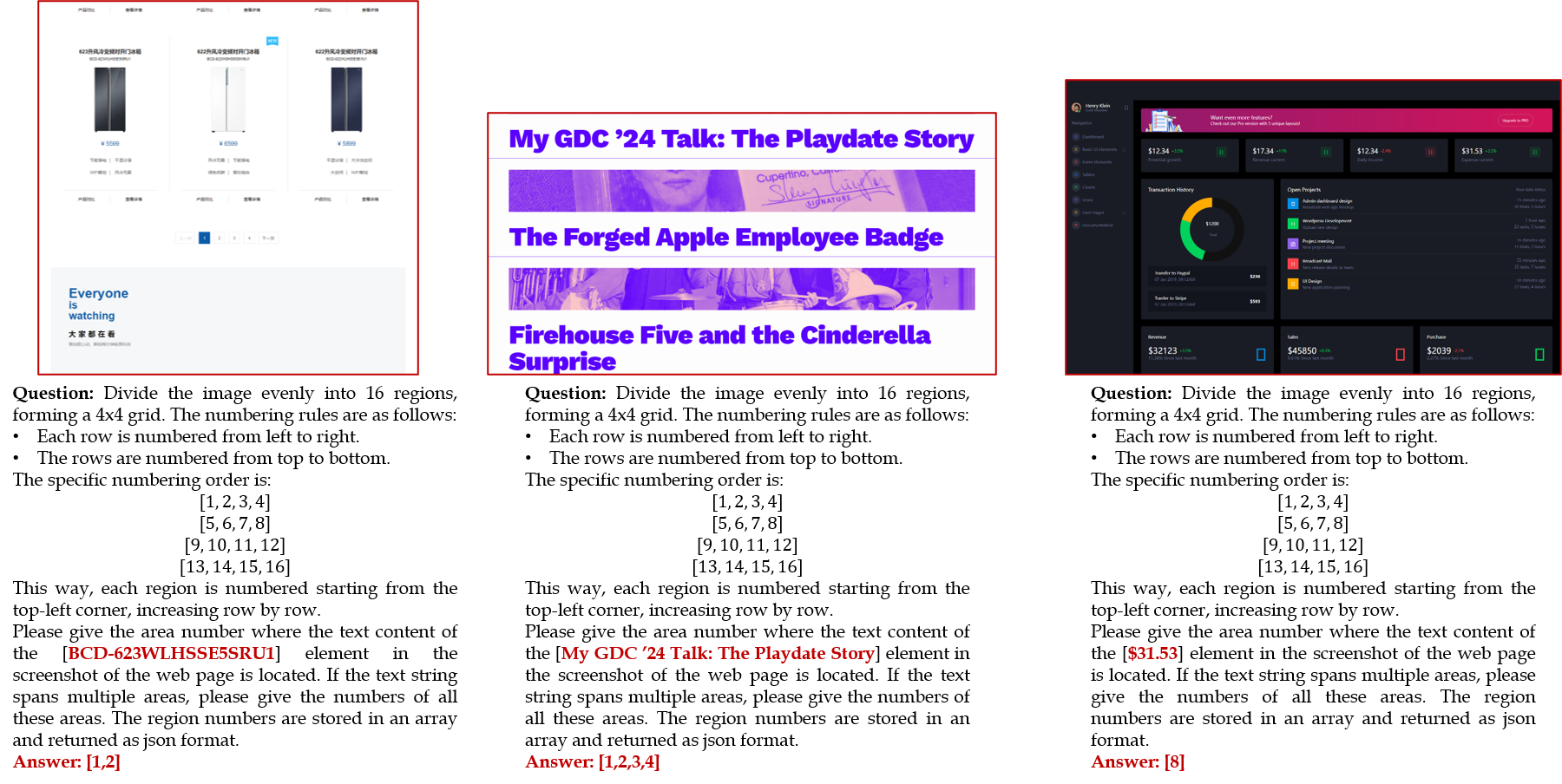}
    \caption{Samples of Visual Grounding(coarse granularity).}
    \label{fig:vg-qa}
\end{figure*}

\begin{figure*}
    \centering
    \includegraphics[width=1.0\linewidth]{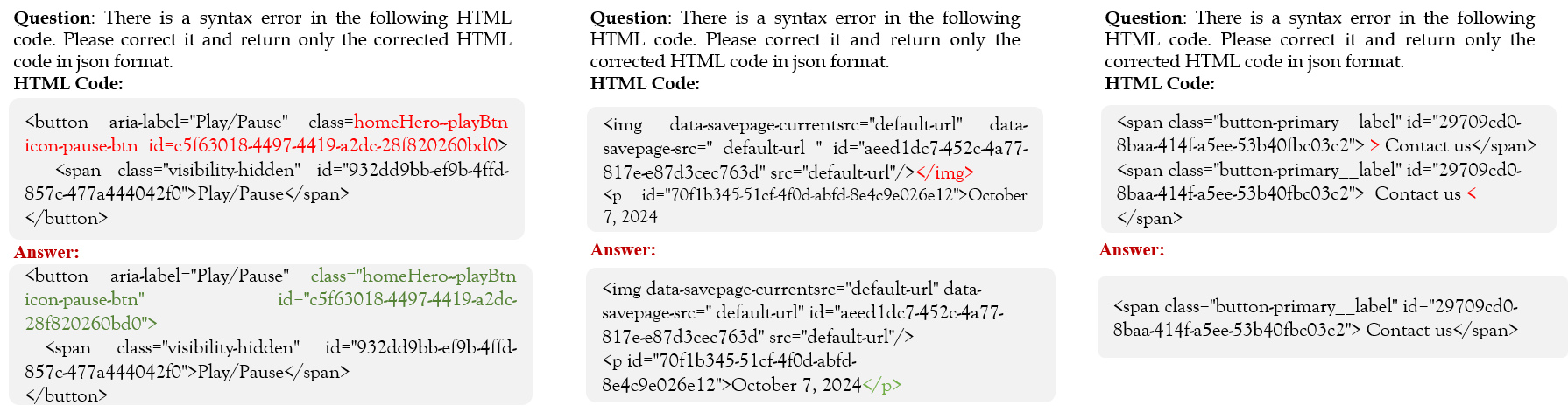}
    \caption{Samples of Code Error Correction.}
    \label{fig:vg-qa}
\end{figure*}

\begin{figure*}
    \centering
    \includegraphics[width=1.0\linewidth]{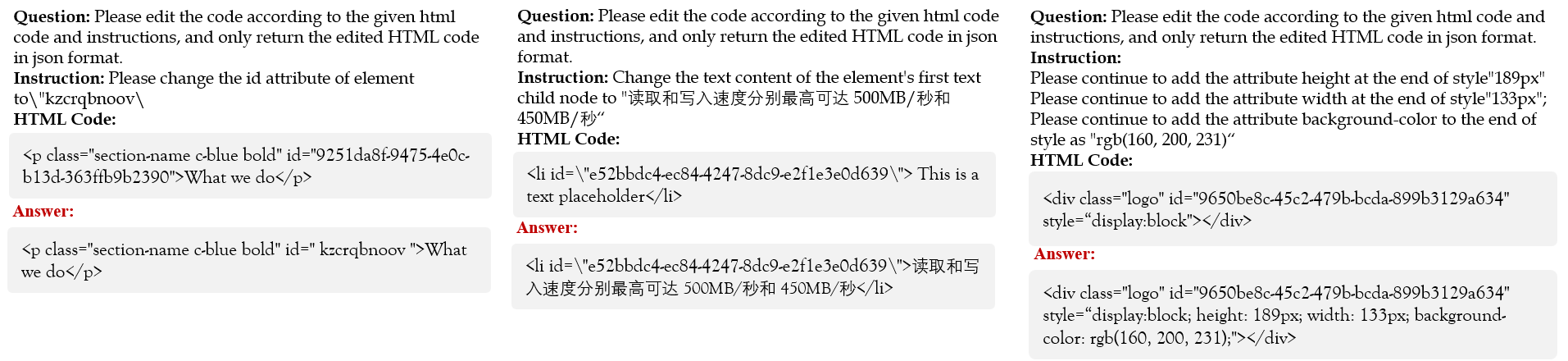}
    \caption{Samples of Code Function Editing.}
    \label{fig:task7}
\end{figure*}

\subsection{Data Collection}
\label{apend:sec2}
\noindent \textbf{Webpage Simplification Algorithm:} In the collected dataset of raw web pages, some pages are overly long, particularly those from news or e-commerce sites. These pages surpass the maximum input length of current multimodal large models, creating challenges for subsequent evaluations. Additionally, these pages contain numerous redundant and duplicate elements. To address this, we developed a web page simplification algorithm that analyzes the DOM tree structure to identify and remove redundant nodes, thereby facilitating more effective evaluations. The specific process of the web page simplification algorithm is shown in Alg \ref{alg:alg2}.

\vspace{2mm}

\noindent \textbf{Webpage Segmentation Algorithm:} The web page segmentation algorithm is designed to overcome the input length limitations of large models. By dividing web pages into multiple slices, we can incrementally process and analyze web content, ensuring each slice is manageable by the model. This approach not only enhances the model's processing efficiency but also improves evaluation accuracy. The process of the web page segmentation algorithm is shown in Alg \ref{alg:alg1}.

\subsection{Automatic Labeling Strategy}
\label{apend:sec3}
\paragraph{Task 1: Element Classification}
We first use the front-end code to determine whether the following three types of tags and their combinations exist on the web page: \textbf{<input />, <button />, <img />}. We then construct them into correct options, such as \textbf{B. <img /> and <input />}. At the same time, we construct error interference options by permuting and combining elements that do not exist on the page; for example: \textbf{A. <img /> and <button />, C. <input /> and <button />, D. None of the above.}

\paragraph{Task 2: Attribute Perception}
To prevent interference in the testing process from elements with identical text content, we initially selected web page elements with unique text. Using the element IDs, we extracted the following four types of style content from the CSS file to create question-answer pairs.
\begin{itemize}
    \item \textbf{background-color} We construct questions and correct answers based on the RGB color format requirements, such as \textbf{rgb(19,25,36)}.
    \item \textbf{color} We construct questions and correct answers based on the RGB color format requirements, such as \textbf{rgb(19,25,36)}.
    \item \textbf{font-style} We construct the following multiple-choice questions: \textbf{A. Italic, B. Oblique.}
    \item \textbf{border-radius} We construct the following multiple-choice questions: \textbf{A. Rounded corners, B. Square corners.}
\end{itemize}

\paragraph{Task 3: Visual Grounding  in the Webpage}
To prevent interference in the testing process from elements with identical text content, we first selected web page elements with unique text content. Next, we retrieved the spatial position information of these elements based on their IDs: \([x1, y1, x2, y2]\). Using this spatial information, we constructed two types of visual grounding tasks. For the grid localization task, we divided the webpage screenshot into a 4x4 grid and automatically calculated the grid numbers occupied by the elements based on their coordinates.

\paragraph{Task4: OCR in the Webpage}
To prevent interference in the testing process from elements with identical text content, we first selected web page elements with unique text. Next, we sorted all text by length and chose short, medium, and long texts to construct question-answer pairs. Finally, using the elements' coordinate information, we drew red borders on the corresponding webpage screenshots to guide the model in performing OCR tasks.

\paragraph{Task 5: Code Error Correction}
To construct correction samples, we designed various code corruption methods based on real web element code to ensure diversity in error types, comprehensively covering common front-end code errors. The specific methods are as follows:
\begin{itemize}
    \item \textbf{Missing Closing Tag.}
    \textit{Description:} Missing a closing tag, resulting in incomplete web elements.
    \textit{Method:} Delete the closing tag of real web elements to generate error samples.
    
    \item  \textbf{Incorrect Character Escaping.}
    \textit{Description:} Certain characters need escaping; otherwise, they interfere with HTML parsing.
    \textit{Method:} Insert random special symbols such as \&, <, > into element text to create escaping errors.
    \item  \textbf{Tag Spelling Error.}
    \textit{Description:} Tag name spelling error, such as writing <p> as <p1>.
    \textit{Method:} Randomly modify the end tag by adding erroneous characters.
    \item \textbf{Attribute Syntax Error.} \textit{Description:} Attribute values not enclosed in quotes, causing syntax errors.
    \textit{Method:} Remove quotes from page element attribute values.
    \item \textbf{Attribute Spelling Error.}
    \textit{Description:} Attribute name spelling error, such as writing class as clbss, possibly causing style loss.
    \textit{Method:} Replace some attribute names of elements with misspelled versions.
    \item \textbf{Erroneous Addition of Tags.} 
    \textit{Description:} Adding a closing tag to tags that do not require one (\textit{e.g.,} <img />, <input />).
    \textit{Method:} Add erroneous "closing tags" to these tag types.
    
\end{itemize}






For each real web element, one of the above methods is randomly selected with equal probability to corrupt the code, producing erroneous element code, with the original web code serving as ground truth. This approach generates a large and diverse set of correction samples, aiding in a comprehensive evaluation of MLLM's ability to correct various front-end code errors.

\paragraph{Task 6: Code Function Editing.}

Similarly, based on real web elements, we construct code editing instructions and edited web element code. To cover various common code editing scenarios, we have also designed multiple editing methods. The specific editing methods are described as follows:

\begin{itemize}
    \item \textbf{Modify Element Attributes.}
    \textit{(i)} Randomly select an existing attribute of an element to modify, such as class, id, href, etc, and\textit{(ii)} Generate random values for the selected attribute and replace it.
    \item \textbf{Modify Element Text.}
    \textit{Uniformly modify the text content of elements (if any) to a placeholder:} Replace the first text child node of the element (if any) with "This is a text placeholder."
    \item \textbf{Add or Modify Element Style.}
    \textit{Modify or add style attributes of elements, such as height and background-color:} If the attribute exists, modify its value; otherwise, append a new style definition in the style attribute.
    \item \textbf{Add or Delete Child Nodes.}
    \textit{(i)} Randomly delete a child node (if any) or \textit{(ii)} Insert a new child node into the element, randomly set the tag and attributes of the child node, and set its text content to "this is a new node."
\end{itemize}


\begin{table*}[t]
\footnotesize
\centering
\caption{Evalutaion results of WebUI-to-Code at element and layout level. 
Dimensions name: CCSR=Code Compile Success Rate, TS=Text Similarity, CS=Color Similarity, BCS=Background Color Similarity, CGE=Coarse-Grained Evaluation}

\scalebox{1.0}{
\begin{tabular}{lc|cccccc}
\toprule

Model & Size & CCSR & TS & CS & BCS & Layout & CGE   \\ \hline
\multicolumn{8}{l}{\cellcolor[HTML]{c0c0c0}\textit{\textbf{Closed Source Model}}} \\
GPT-4o & - & 95.8 & 73.6 & 72.1 & 81.1 & 89.2 & 81.7  \\
GPT-4o-mini & - & 99.0 & 59.8 & 53.3 & 66.7 & 86.4 & 76.8 \\
Cluad-3.5-Sonnet & - & 85.2 & 72.9 & 70.4 & 79.6 & 88.7 & 81.1 \\
Gemini-1.5-pro  & - & 96.7  & 73.7 & 67.9 & 80.5 & 87.8 & 78.8  \\
Yi-Vision & - & 96.1 & 61.4 & 61.1 & 72.8 & 86.5 & 78.4 \\
GLM-4v & -  & 77.5 & 58.9 & 55.6 & 68.4 & 85.5 & 74.8  \\ 
Step-1.5v-mini & - & 62.3 & 51.4 & 59.1 & 73.5 & 85.5 & 75.4 \\
\hline
\multicolumn{8}{l}{\cellcolor[HTML]{c0c0c0}\textit{\textbf{Open Source Model}}} \\
Qwen2-VL & 2B & 11.8 & 51.8 & 54.2 &  79.7 & 83.8 & 68.0 \\
InternVL2 & 2B & 4.0 & 48.5 & 29.7 & 60.3 & 80.9 & 67.0 \\
InternVL2.5 & 2B & 18.1 & 47.9 & 39.3 & 55.9 & 84.0 & 62.7 \\
Ovis1.6-Llama3.2 & 3B & 62.4 & 57.5 & 45.7 & 67.5 & 83.2 & 68.6 \\
InternVL2 & 4B & 38.2 & 56.6 & 47.1 & 64.9  & 83.2 & 68.6 \\
InternVL2.5 & 4B & 18.7 & 50.8 & 46.6 & 64.5 & 84.7 & 74.2 \\ \midrule
Qwen2-VL & 7B & 47.3 & 54.1 & 49.2 & 64.3 & 84.6 & 71.8 \\
InternVL2 & 8B & 81.5 & 62.8 & 51.3 & 70.7& 84.1 & 71.2 \\
InternVL2.5 & 8B & 95.3 & 58.8 & 49.7 & 63.5& 84.3 & 73.4 \\
MiniCPM-V-2.6 & 8B & 90.9 & 58.0 & 46.6 & 64.7 & 85.0 & 71.9 \\
Phi-3-vision & 8B & 60.5 & 23.7 & 29.8 & 41.8& 82.1 & 64.5 \\
Phi-3.5-vision & 8B & 53.6 & 21.7 & 27.3 & 38.6& 81.1 & 62.5 \\
Ovis1.6-Gemma2 & 9B & 60.3 & 58.5 & 52.5 & 69.2& 85.8 & 74.4 \\ \midrule
InternVL2 & 26B & 75.0 & 57.7 & 48.3 & 64.0 & 84.3 & 69.4 \\
InternVL2.5 & 26B & 91.6 & 58.9 & 63.5 & 69.1& 84.9 & 76.9 \\
Ovsi1.6-Gemma2 & 27B & 80.8 & 62.7 & 58.4 & 72.4& 85.7 & 76.1 \\
InternVL2.5 & 38B & 86.6 & 59.2 & 57.9 & 72.7& 87.3 & 76.5 \\
InternVL2 & 40B & 84.6 & 67.5 & 57.7 & 76.7& 85.9 & 74.1 \\ \midrule
Qwen2-VL & 72B & 83.7 & 69.5 & 61.5 & 74.8& 88.2 & 79.2\\
NVLM-D & 72B & 23.7 & 52.8 & 45.8 & 70.1& 83.9 & 69.4 \\
InternVL2 & 76B & 94.9 & 63.3 & 55.7 & 71.7& 86.6 & 75.3 \\
InternVL2.5 & 78B & 85.4 & 65.6 & 59.8 & 74.5& 86.9 & 77.0 \\ \bottomrule
\label{tab:w2cfine}
\end{tabular}}
\end{table*}

\begin{figure*}[ht]
    \centering
    \includegraphics[width=1.0\linewidth]{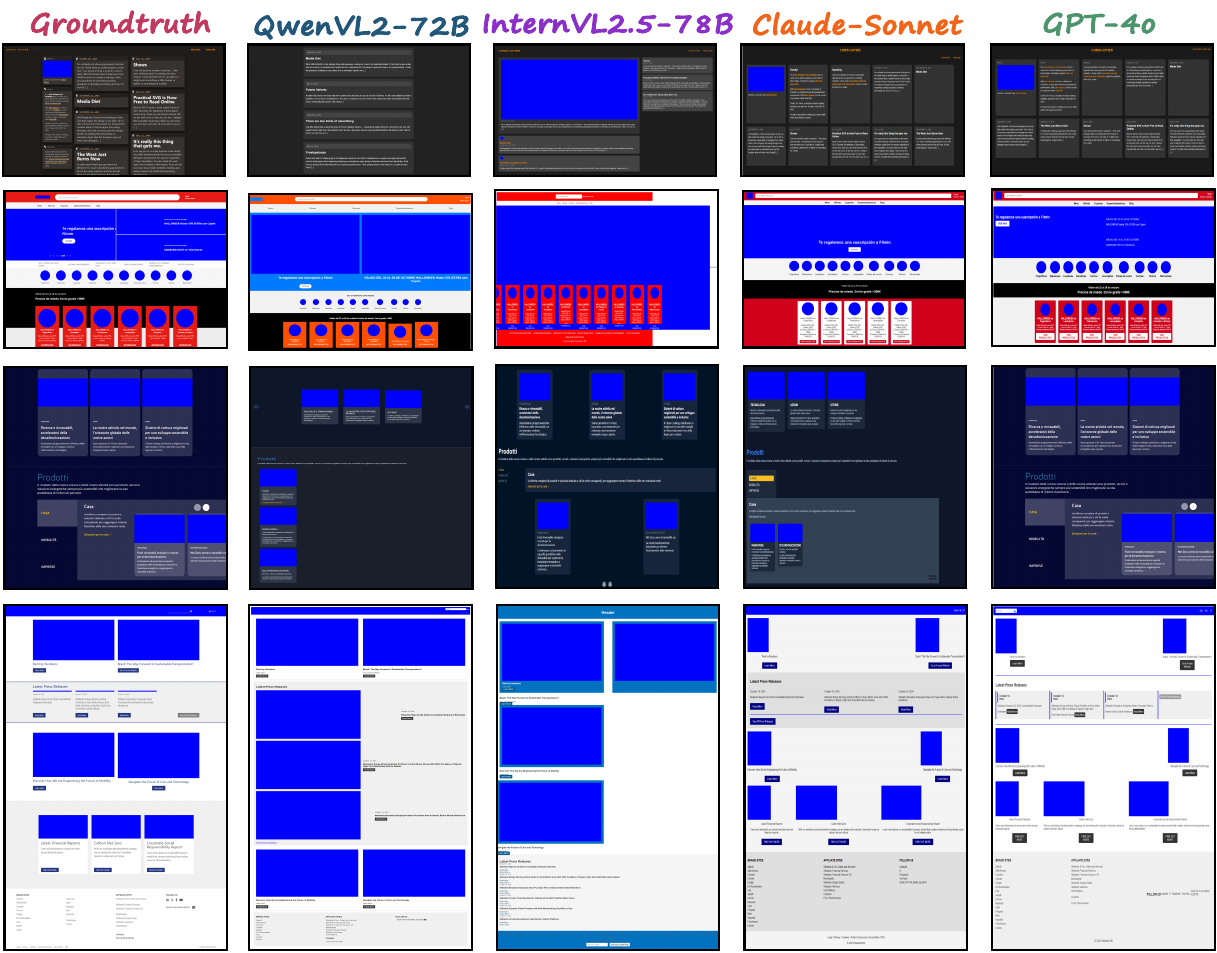}
    \caption{Comparison of webpage generation effects by different models on complex webpages and slices.}
    \label{fig:task10-exa}
\end{figure*}

For each real web element, one of the above editing methods is randomly selected with equal probability to edit the original code, resulting in edited code (as ground truth) and specific editing methods (as editing instructions for MLLM). This approach generates a large and diverse set of code editing samples, aiding in a comprehensive evaluation of MLLM's ability to modify web element code based on diverse editing instructions.

\paragraph{Task 7: Webpage-HTML Matching}
We first matched the corresponding HTML code snippets using the ID set of all elements in the webpage slices. Next, we randomly chose, with equal probability, whether to alter the correct code snippets. For the modification, we selected the code of an element from another slice of the same webpage to replace one element's code in the correct snippet.

\paragraph{Task 8: Webpage-HTML Retrieval}
We first matched the corresponding HTML code using the ID set of all elements in the webpage slices and used it as the correct option. For the incorrect options, we randomly selected elements and their corresponding code snippets from other slices on the same website, replacing parts of the correct option's code at different levels. We chose to replace 1, 2, or 3 elements to construct three incorrect options.

\subsection{More Results}
We show more detailed experimental results of all models at the element level and layout level in Table \ref{tab:w2cfine}. At the element level, we supplement the evaluation results at different fine-grained dimensions, including text content, font color and background color.

Figure \ref{fig:task10-exa} shows the generation effects of more models on webpage slices and full webpages. We selected the top-2 models of open source and closed source models for display. It can be seen that as the complexity of web page content increases, the generation effect of the model gradually deteriorates.

\begin{algorithm*}
\caption{Simplify DOM Tree}
\KwIn{Original DOM tree $T$;similarity threshold $\delta$}
\KwOut{Simplified DOM tree $T'$}

\Begin{
    Initialize $T'$ as a copy of $T$\;
    Group child nodes by TagName\;
    \ForEach{group of child nodes}{
        \ForEach{pair of nodes $(n_1, n_2)$ in the group}{
            Split ClassName of $n_1$ and $n_2$ into list $\mathcal{S}_1$ and $\mathcal{S}_2$\;
            Calculate Dice similarity:
            $
            \text{Dice} \leftarrow \frac{2 \times |\mathcal{S}_1 \cap \mathcal{S}_2|}{|\mathcal{S}_1| + |\mathcal{S}_2|}
            $
            
            \If{Dice similarity $\delta$}{
                Group $n_1$ and $n_2$ together\;
            }
        }
        \If{group size $\leq 5$ \textbf{or} all nodes have similar heights}{
            Retain all nodes in the group\;
        }
        \Else{
            Sort nodes by bottom height in ascending order and retain the top 50\%\;
        }
    }
}
\Return{Simplified DOM tree $T'$}
\label{alg:alg1}
\end{algorithm*}

\begin{algorithm*}
\caption{Webpage Slices Generation}
\KwIn{Webpage Screenshots, slice height $H_s$; minimum slice height $H_{min}$; webpage height $H_{page}$}; Element coordination set $\mathcal{S}\leftarrow\{[x_1^1,y_1^1,x_2^1,y_2^1],...[x_1^n,y_1^n,x_2^n,y_2^n]\}$

\KwOut{List of webpage slices}

\Begin{
    
    
    Initialize slice top boundary $H_{t}\leftarrow0$\ and slice bottom boundary $H_b\leftarrow H_s$;

    \While{ $H_b<H_{page}$}{
        Filter the elements which  $y_2>H_b$ and $y_1<H_b$\;
        $y_{max}\leftarrow$ calculate the maximum value of set $y_2$\;
        
        \If{$H_{page}-y_{max} \leq H_{min}$}{
            $y_{max} \leftarrow H_{page}$\;
        }
        
        Retrieve element IDs and save webpage slice between $H_t$ and $y_{max}$\;

        $H_t \leftarrow y_{max}$\;
        $H_b \leftarrow y_{max}+H_s$\;

        \If{$H_b > H_{page}$}{
            Retrieve element IDs and save webpage slice between $H_t$ and $H_{page}$\;
            \textbf{break}\;
        }
        
    }
    
    \Return{All webpage slices, element IDs}\;
}
\label{alg:alg2}

\end{algorithm*}

\end{document}